\documentclass[11pt]{article}

\usepackage[final]{acl}

\usepackage{times}
\usepackage{latexsym}
\usepackage[utf8]{inputenc}
\usepackage[T1]{fontenc}
\usepackage{microtype}
\usepackage{graphicx}
\usepackage{subcaption}
\usepackage{booktabs}
\usepackage{xcolor}
\usepackage[normalem]{ulem}
\usepackage{xspace}
\usepackage{multirow}
\usepackage{amsmath}
\usepackage{amssymb}
\usepackage{mathtools}
\usepackage{amsthm}
\usepackage{nicefrac}
\usepackage{tabularx}
\usepackage{bm}
\usepackage{array}
\usepackage{pifont}
\usepackage{algorithm}
\usepackage{algorithmic}
\usepackage{float}
\usepackage[capitalize,noabbrev]{cleveref}
\usepackage[most]{tcolorbox}

\newcolumntype{C}[1]{>{\centering\arraybackslash}m{#1}}

\theoremstyle{plain}

\theoremstyle{definition}

\theoremstyle{remark}

\newcommand{\apebKeepD}[1]{#1}

\newcommand{\apebKeepC}[1]{#1}
\newcommand{\apebKeepE}[1]{#1}
\newcommand{\apebKeepB}[1]{#1}
\newcommand{\apebKeepA}[1]{#1}
\newcommand{\apebKeepF}[1]{#1}

\newcommand{\apebKeepH}[1]{\textcolor{black}{#1}}

\newcommand{\apebKeepJ}[1]{\textcolor{black}{#1}}
\newcommand{\apebKeepK}[1]{\textcolor{black}{#1}}

\newcommand{\benchf}{\textbf{APeB}\xspace}
\newcommand{\benchfull}{\textbf{A}gent \textbf{Pe}rsonalized \textbf{B}enchmark\xspace}
\newcommand{\benchs}{\text{APeB}\xspace}

\newcommand{\smartq}{\text{intent}\xspace}
\newcommand{\smartqB}{\text{Intent}\xspace}
\newcommand{\matchq}{\text{refined}\xspace}
\newcommand{\matchqB}{\text{Refined}\xspace}
\newcommand{\ouragent}{\text{VQRA}\xspace}
\newcommand{\ouragentB}{\text{VQRA}\xspace}
\newcommand{\ouragentf}{\text{V}ague \text{Q}uery \text{R}efinement \text{A}gent\xspace}

\newenvironment{myquotation}{\setlength{\leftmargini}{0em}\quotation}{\endquotation}

\newtcolorbox{expfindings}{
  colback=yellow!10,
  colframe=black!50,
  title=Key Experimental Findings,
  fonttitle=\bfseries,
  sharp corners
}

\newtcolorbox{prompt}{
  enhanced,
  breakable,
  colback=white!10,
  colframe=black!100,
  title=Prompt,
  fonttitle=\bfseries,
  sharp corners
}

\newcommand{\emailauthor}[2]{\href{mailto:#1}{\textcolor{black}{\textbf{#2}}}}

\title{APeB: Benchmarking Personalization Ability of \\ Large Language Model Agents}

\author{
  \emailauthor{hcyang@cse.cuhk.edu.hk}{Garry Yang\textsuperscript{1,*}},
  \emailauthor{zzchen2@cse.cuhk.edu.hk}{Zizhe Chen\textsuperscript{1,*}},
  \emailauthor{chenxinru.xr@bytedance.com}{Xinru Chen\textsuperscript{2}},
  \emailauthor{yqchen@cse.cuhk.edu.hk}{Yongqiang Chen\textsuperscript{1}}
\\
  \emailauthor{wangjianxiang.4181@bytedance.com}{Jianxiang Wang\textsuperscript{2}},
  \emailauthor{dyzou24@cse.cuhk.edu.hk}{Deyu Zou\textsuperscript{1}},
  \emailauthor{linyi.ding@bytedance.com}{Linyi Ding\textsuperscript{2}},
  \emailauthor{wujialiang.1010@bytedance.com}{Jialiang Wu\textsuperscript{2}}
\\
  \emailauthor{yunzhong.he@bytedance.com}{Yunzhong He\textsuperscript{2}},
  \emailauthor{gy910210@gmail.com}{Yu Gong\textsuperscript{2}},
  \emailauthor{jcheng@cse.cuhk.edu.hk}{James Cheng\textsuperscript{1,\textdagger}},
  \emailauthor{zhangyuan.zhang@bytedance.com}{Huaixiao Tou\textsuperscript{2,\textdagger}}
\\[0.5ex]
  {\normalfont\textsuperscript{1}The Chinese University of Hong Kong
  \quad
  \textsuperscript{2}ByteDance}
\\
  \apebKeepH{\texttt{hcyang@cse.cuhk.edu.hk}}
}

\begin{document}

\maketitle
\begingroup
\renewcommand{\thefootnote}{\fnsymbol{footnote}}
\footnotetext[1]{Equal contribution.}
\footnotetext[2]{Corresponding authors.}
\endgroup

\begin{abstract}
LLM-powered agents struggle with personalization when users issue raw, underspecified queries. In this setting, agents must infer latent intent, extract preferences from noisy interaction histories, and select among competing alternatives. Existing benchmarks rarely test this capability, as they often rely on user-refined queries or simplified histories.
We introduce personalized product search (PPS), a testbed for agentic personalization under raw queries and diverse histories. We construct \benchfull (\benchf) from action logs, pairing underspecified intents with rich histories and user-viewed candidate items. Evaluating state-of-the-art LLMs with multi-step agent workflows, we find that models handle explicit queries well but struggle with early-stage queries requiring intent and preference discovery. 
Rubric analysis attributes this gap mainly to ineffective history use. 
A simple history-aware query-refinement pipeline, \ouragent, yields consistent gains, highlighting the need for dedicated history-utilization modules in personalized agents.
\end{abstract}

\section{Introduction}
\begin{figure}[t]
    \centering
    \includegraphics[width=\columnwidth]{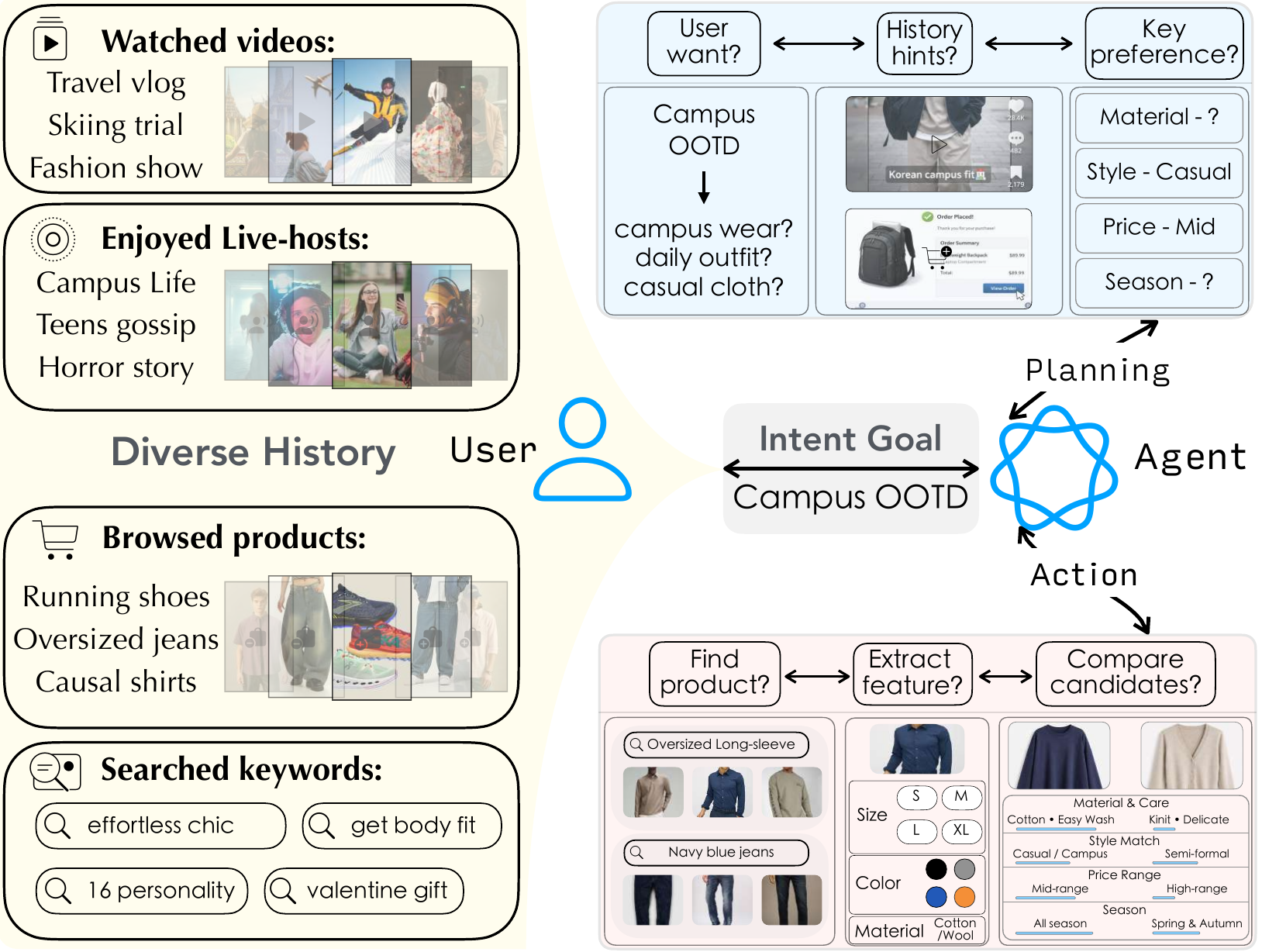}
    \caption{\apebKeepA{Personalization challenge for agents: infer intent and preferences from diverse history to make a preference-aligned decision.}}
    \label{fig:introduction}
\end{figure}

\begin{table*}[t]
\centering
\resizebox{\textwidth}{!}{%
\begin{tabular}{lllll}
\toprule
\textbf{Category} & \textbf{Persona} & \textbf{User Goal} & \textbf{Candidate Set} & \textbf{Evaluation} \\
\midrule
Generic Shopping & -- & Explicit, multi-constraint & Web-scale / open-ended & Multi-step web shopping \\
\midrule
Recommendation & Homogeneous, low-semantic & -- & Open-ended & Next-item prediction \\
\midrule
Personalized Product Search & Homogeneous, low-semantic & Explicit, user-refined & Randomly sampled & Product finding \\
\midrule
\apebKeepA{Rationale} Prediction & Synthetic, surveys & -- & Random / open-ended & Localized reasoning \\
\midrule
Multi-turn Memory & Homogeneous, synthetic & Explicit, instruction-driven & -- & Memory-aware generation \\
\midrule
\textbf{\benchf} & \textbf{Heterogeneous, latent} & \textbf{Implicit, ambiguous} & \textbf{Few closely competing} & \textbf{Personalized goal completion} \\
\bottomrule
\end{tabular}
}
\caption{\apebKeepA{Comparison of existing benchmarks by whether and how they capture core personalization challenges.}}
\label{tab:compare}
\end{table*}

Large Language Model (LLM)-powered agents extend foundation models with additional modules for memory, planning, and execution, enabling multi-step, goal-directed reasoning over long contexts 
\cite{memory-survey-2, memory-survey-1, memory-survey-5, MemoryAgentBench}.
As shown in Figure \ref{fig:introduction}, a central challenge for LLM-powered agents is \emph{personalization}: 
\apebKeepD{user objectives are not explicitly specified; preferences are implicit and partially observed through histories; and correctness is user-centric rather than globally optimal \cite{personalization-underground-2}.}
In practice, users implicitly reason over their own past experience, translate vague intent into specific criteria, and discriminate between plausible options \cite{switch-query-3, long-trail-1, query-dataset-3-JDsearch}. 
\apebKeepJ{Personalized decision-making requires agents to align their planning and actions with a user's intent and preferences~\cite{memory-survey-5}. To study this capability when user goals are not fully specified, we ask:}
\begin{myquotation}
\apebKeepD{\emph{How can we benchmark the agentic personalization capability of LLMs?}}
\end{myquotation}
\apebKeepA{To our knowledge, existing benchmarks individually address one or two of these axes but rarely combine noisy heterogeneous histories, vague-to-refined search trajectories, and closely competing candidates.}
\emph{Recommendation benchmarks} cast personalization as next action prediction from historical signals, emphasizing preference deduction over objective understanding \cite{taobao-dataset, repeat-buy}. 
\emph{Rationale prediction benchmarks} evaluate localized reasoning, focusing on generating intermediate justifications in isolation rather than reaching a user-aligned outcome \cite{opera, SessionIntentBench}.
\emph{Multi-turn Memory benchmarks} model personalization over largely homogeneous conversations, evaluating relevant information extraction instead of intent inference under ambiguity \cite{MemoryAgentBench, Mem-PAL}.
In all, these benchmarks sidestep the need to jointly reason over noisy histories, unrefined user queries, and closely competing candidates.
\apebKeepJ{As a result, they} do not directly evaluate whether agentic pipelines (e.g., ReAct) contribute to intent inference and preference-aligned actions \cite{agent-adv-1}.

Motivated by \apebKeepA{these limitations}, we construct \benchf (\benchfull) to evaluate agents' personalization ability through \apebKeepA{the} \emph{Personalized Product Search} (\textbf{PPS}) task.
PPS frames personalization as a user-specific choice among competing products conditioned on a shopping intent and long-term behavioral preference \cite{ps-3, ps-2}.
On online platforms, user behavior is noisy across domains, shopping intent is often \apebKeepE{vague and early-stage}, and purchases occur only after comparing similar products \cite{onerecthink, query-dataset-1, switch-query-3, hard-neg-1}.
However, prior \emph{PPS benchmarks} rely on sanitized histories, clarified queries, and easily separable candidates \cite{query-dataset-3-JDsearch, amazon-review1},
\apebKeepD{thereby failing to evaluate personalization under ambiguous, reasoning-intensive conditions.} 
\apebKeepA{The preference-aligned purchase also contrasts with \emph{generic shopping benchmarks} that focus on isolated sub-tasks (e.g., product matching) without grounding decisions in user-specific intents and histories \cite{MMLU, WebShop}.}
\apebKeepH{Table~\ref{tab:compare} summarizes this comparison.}

In turn, \benchf contains intricate cases drawn from a content platform with shopping functionality.
\apebKeepC{It includes 5,648 cases from 10+ categories
\apebKeepH{sampled over a 30-day window, with up to 60 days of history preceding the first query of each retained session 
(Section~\ref{sec:bench-dis})}.}
\apebKeepE{Users primarily consume videos and livestreams, yet can seamlessly transition to product search, browsing, and purchase within the same platform. This end-to-end flow yields rich, heterogeneous interaction histories that capture real-world personalization challenges \apebKeepC{(see Appendix~\ref{ap-benchmark-platform})}.}
We disentangle user behavior into a long-term history period and a personalized shopping session, delimited by the timestamp of a user-issued search (Section \ref{sec:bench-dis}).
Sessions are retained only if the search leads to a query-aligned purchase following reasoning-intensive browsing (Section \ref{sec:bench-filter}).
The history period comprises diverse interactions with products and contextual content (e.g., videos and livestreams), providing long-term personalization signals (Section \ref{sec:bench-history}). The shopping session contains user searches spanning vanilla intent to refined keywords (Section \ref{sec:bench-query}), together with a set of heavily viewed competitive candidates (Section \ref{sec:bench-candidate}) and a \apebKeepA{final observed purchase}, enabling intent evaluation and verifiable user-centric assessment.
\apebKeepA{We evaluate whether personalization agents} can first infer a user's latent preferences from vague intents and history hints (\emph{Planning}) and select among close alternatives (\emph{Action}) on the user's behalf \apebKeepA{(the pipeline in Figure \ref{fig:introduction})}.

We test \apebKeepA{latest LLMs} with agentic workflows on \benchs.
Evaluation spans \smartq and \matchq queries with random and hard candidates, and includes comparison against a PPS model UniSAR \cite{pps-s-6}.
Results show that (1) \apebKeepJ{with either well-specified \matchq queries or random candidate sets}, LLMs substantially outperform UniSAR, and ReAct workflows further improve performance; (2) \apebKeepJ{when vague, early-stage \smartq queries are paired with hard candidate sets}, LLMs fall to the level of trained UniSAR, with limited gains from ReAct workflows;
(3) rubric analysis shows that \smartq queries expose weak intent and history attribution, \apebKeepK{with ReAct traces showing lower reasoning coherence and history-retrieval utility;} (4) 
\apebKeepA{we find that a simple history-grounded \ouragentf (\ouragentB) pipeline improves intent inference and Hit@1 over single-prompt under \smartq.
}
\apebKeepJ{Overall, \benchs provides a behavior-grounded testbed for evaluating how agentic systems use long-term histories to infer and fulfill underspecified user goals.}

\section{\apebKeepH{Related Work}}\label{sec:related}
\paragraph{\apebKeepH{LLM-powered personalization agents.}}
\apebKeepK{LLM-powered personalization agents use user models and long-term memories to condition current decisions on past interactions~\cite{memory-survey-1,MemGPT,o-mem}.
User representations range from explicit profiles to latent personas~\cite{lamp,persona-gen-1,persona-gen-2,persona-gen-3,Mem-PAL}; memory systems store and retrieve task-relevant evidence from past interactions~\cite{longmem}; and personalized search methods use history-derived preference signals for query-aware refinement and product reranking~\cite{memrerank}.
Such systems must identify which interactions remain relevant as preferences and intent evolve, then ground the resulting evidence in fine-grained candidate comparison.}
\apebKeepJ{Resolving vague intents requires coordinating history retrieval, preference inference, query formulation, and candidate comparison~\cite{o-mem,Harnessing,ReAct}. 
Multi-step LLM agents can struggle with intermediate decisions~\cite{zou2025reducing,zou2026information,chen2026histanumcaestimatestate}; in personalization, this may surface as failures to identify or apply task-relevant preferences.
}

\paragraph{\apebKeepH{Generic e-commerce benchmarks.}}
\apebKeepH{E-commerce benchmarks cover product search, web navigation, and shopping knowledge. Evaluation spans retrieval accuracy, task success, generation quality, and rubric scores~\cite{query-dataset-1,eCeLLM}. Search datasets center on direct queries or synthetically constructed multi-constraint requests~\cite{amazon-review1,query-dataset-3-JDsearch,query-dataset-2-DeepShop,query-dataset-6-ShoppingComp}. Web-agent settings add interactive navigation and product comparison, often through synthetic or annotator-authored goals~\cite{WebArena,WebShop,mind2web,shoppingbench}. Knowledge benchmarks decompose shopping into QA-style concept and attribute reasoning~\cite{eCeLLM,EcomEval,MMLU}. These settings primarily evaluate explicit task completion or isolated reasoning components.}

\paragraph{\apebKeepH{Behavior datasets for personalization.}}
\apebKeepH{Recommendation datasets learn from user--item histories through next-item or rating prediction~\cite{amazon-m2,repeat-buy,taobao-dataset,yoochoose}. PPS further combines preference and query intent, progressing from user--query--item representations to query-aware and LLM-based retrieval~\cite{pps-1,pps-2,pps-3,pps-s-3,pps-s-6,rs-llm-1}. Semantic and LLM-based variants improve textual generalization, while jointly using long histories, vague queries, and fine-grained item attributes remains challenging~\cite{pps-s-1,pps-s-2,rs-survey-1,rs-survey-3}. Existing datasets commonly pair observed targets with unobserved-item negatives~\cite{query-dataset-3-JDsearch,amazon-review1}, while memory benchmarks test retention of user-specific evidence across synthetic dialogues or sessions~\cite{Mem-PAL,MemoryAgentBench,AgentRecBench,AgentSociety}. APeB joins heterogeneous histories and user-issued vague queries with viewed hard candidates and purchase-linked evaluation, connecting history-grounded reasoning to the user's realized choice.}

\section{\benchf Construction}\label{sec:bench}

\begin{figure*}[t]
    \centering
    \includegraphics[width=\textwidth]{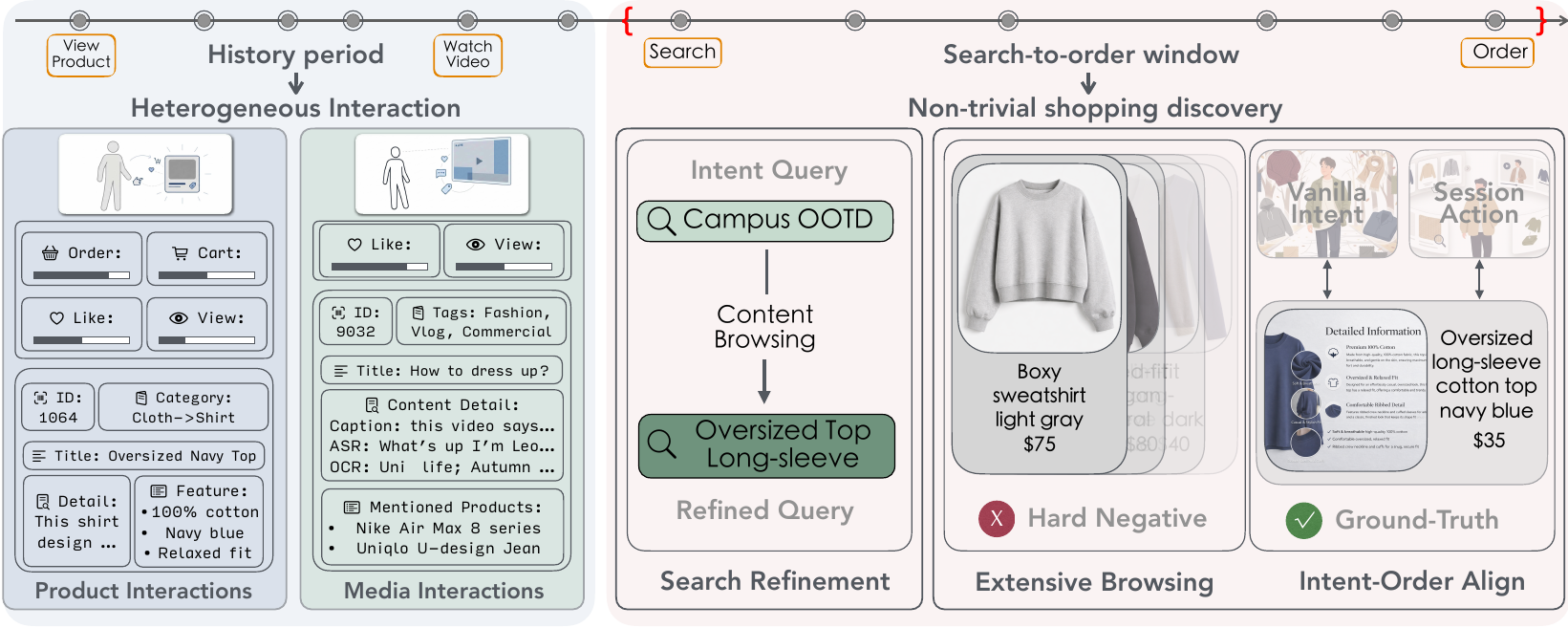}%
    \caption{Construction Pipeline of \benchf:
    from behavioral logs to benchmark components:
    curated heterogeneous interactions, non-trivial shopping session discovery
    (search refinement, extensive browsing, and intent--order alignment),
    and \smartq/\matchq query--candidate pair identification. 
    \apebKeepD{\benchs targets challenging personalization grounded in holistic user behavior.}}
    \label{fig:apeb}
\end{figure*}

We introduce \benchf, a benchmark designed for evaluating \textbf{agentic personalization capability} on the PPS task.
To retain cases in which the personalized decisions are time-consuming and interaction-heavy (Section \ref{sec:bench-filter}), we collect \textbf{non-trivial user shopping sessions} with their heterogeneous histories mined from holistic user behavior logs (Appendix \ref{ap-benchmark-collect}). 
As shown in Figure \ref{fig:apeb}, a \emph{session} is defined as a time-bounded search-to-order window that begins with a search and ends with a purchase (Section~\ref{sec:bench-dis}). 
Non-trivial sessions involve query reformulation and extensive product comparison before an intent-aligned order (Section \ref{sec:bench-filter}), reflecting how user history shapes query formulation, exploration, and decision-making~\cite{AgentRecBench, history-aware-query-1}.
\apebKeepJ{Together, these components are designed to reduce shortcuts based solely on surface-level semantic matching and increase the need for joint reasoning over historical preference signals, query intent, and fine-grained candidate differences.}
For filtered sessions, we construct three benchmark components:
(i) \textbf{Heterogeneous histories}--preference signals (Figure \ref{fig:apeb} Left, Section \ref{sec:bench-history});
(ii) \textbf{User-issued \smartq (vague, early-stage) and \matchq (well-specified, late-stage) queries}--different levels of query specificity (Figure \ref{fig:apeb} Right, Section \ref{sec:bench-query});
and (iii) \textbf{Hard candidate set}--products the user considered, together with the \textbf{
    \apebKeepA{observed purchased item}} (Figure \ref{fig:apeb} Right, Section \ref{sec:bench-candidate}).

\subsection{Non-trivial Shopping Session Discovery}\label{sec:benchmark-discover}
\apebKeepH{We identify non-trivial shopping sessions over a 30-day sampling window. On each day, we first retain users with at least five product or media interactions on every day of the preceding 30 days (Appendix~\ref{ap-benchmark-filter-user}).}
Following the long--short period division paradigm~\cite{repeat-buy}, 
we then extract \textbf{search-to-order sessions} by anchoring boundaries at search timestamps.
From these, we further select \textbf{non-trivial cases} characterized by query evolution, extensive content browsing, and cross-product comparison prior to a personalized purchase decision.

\subsubsection{Search-to-Order Session Disentangle}\label{sec:bench-dis}
We disentangle each user's interaction sequence by the search action timestamp \apebKeepA{(`\{' in Figure \ref{fig:apeb}).}
For a user \(u\), let
\(
\mathcal{I}_u = \{(x_u^k, t_u^k)\}_{k=1}^{N_u}
\)
denote the chronologically ordered interaction log, where \(x_u^k\) is an action and \(t_u^k\) its timestamp.
We consider an initial search action \(x_u^m=\mathrm{query}(q)\) at time \(t_u^m\) and a final purchase action
\(x_u^n=\mathrm{order}(p)\) at time \(t_u^n\).
A \emph{search-to-order session} is identified if the purchase occurs within a fixed time window
\(\Delta=30\) minutes after the search:
\(
t_u^n - t_u^m \le \Delta.
\)
\apebKeepH{Users without any such session are removed. After applying the session-level criteria below, if a user has multiple qualifying sessions across the 30-day sampling window, we retain the earliest one.}
For each retained session, the search time \(t_u^m\) and the purchase time \(t_u^n\) define the disentanglement boundary.
\apebKeepH{We set the pre-session history horizon to \(\Delta_H=60\) days, measured backward from the initial search timestamp \(t_u^m\):}
\begin{align}
    \mathcal{H}_u & = \apebKeepH{\{(x_u^k,t_u^k):t_u^m-\Delta_H\le t_u^k<t_u^m\}}, \label{eq:history}\\
    \mathcal{S}_u & = \{(x_u^k, t_u^k)\in\mathcal{I}_u : t_u^m \le t_u^k \le t_u^n\}. \label{eq:session}
\end{align}
Here, \(\mathcal{H}_u\) captures long-term preference signals, while \(\mathcal{S}_u\) contains goal-driven queries and candidates.
This disentanglement separates preference inference from task execution
, enabling reliable evaluation of personalization reasoning.

\subsubsection{Non-trivial Session Filtering}\label{sec:bench-filter}

For each user \(u\) with a session of ordered events \(\mathcal{S}_u\) (Section~\ref{sec:bench-dis}),
each event is one of
\emph{search}, \emph{product browsing}, \emph{media (video/livestream) browsing}, or \emph{purchase}.
It can be represented as
\[
\mathcal{S}_u = \big( \mathcal{Q}_u^{\mathrm{search}},\ 
          \mathcal{P}_u^{\mathrm{view}},\ 
          \mathcal{V}_u^{\mathrm{view}},\ 
          \mathcal{P}_u^{\mathrm{order}} \big),
\]
where
$\mathcal{Q}_u^{\mathrm{search}}$ is the ordered list of search queries,
$\mathcal{P}_u^{\mathrm{view}}$ the product-interactions,
$\mathcal{V}_u^{\mathrm{view}}$ the media-interactions,
and $\mathcal{P}_u^{\mathrm{order}}$ the purchased products.
To obtain non-trivial shopping sessions, we formulate these three conditions:

\textbf{(1) Iterative query refinement.}
The user progressively clarifies intent through multiple searches:
\begin{equation}\label{eq:query-refine}
    |\mathcal{Q}_u^{\mathrm{search}}| \ge 2.
\end{equation}

\textbf{(2) Extensive-browsing behavior.}
The user exhibits substantial exploration prior to purchase:
\begin{equation}\label{eq:extensive-browsing}
|\mathcal{P}_u^{\mathrm{view}}| + |\mathcal{V}_u^{\mathrm{view}}| \ge \tau_v ,
\end{equation}
\apebKeepK{We set \(\tau_v=10\), requiring at least ten product or media interactions before purchase.}

\textbf{(3) Intent--order alignment.}
\apebKeepK{The purchase must align with a query and have sufficient same-category alternatives.}
Specifically, there exists a purchased product \(p_u^{o} \in \mathcal{P}_u^{\mathrm{order}}\) such that
\begin{equation*}
\begin{aligned}
&\exists\, q \in \apebKeepK{\mathcal{Q}_u^{\mathrm{search}}} \ \text{s.t.}\ \mathbb{I}[\mathrm{Align}(q, p_u^{o})] = 1, \\
&\apebKeepK{\left|\left\{v\in\mathcal{P}_u^{\mathrm{view}}\!\setminus\!\{p_u^o\}:
\mathrm{cat}(v)=\mathrm{cat}(p_u^o)\right\}\right|\ge\tau_a},
\end{aligned}
\end{equation*}
\apebKeepK{We set \(\tau_a=9\) for non-purchased products in the purchase category; all viewed products remain candidates, and one purchase is retained.}

\apebKeepK{Rule-based selection enforces the category floor (Appendix~\ref{ap-benchmark-filter-session-only-rule}); GPT-4.1 verifies query--purchase alignment with human validation (Appendix~\ref{ap-benchmark-filter-session-rule3}).}
\apebKeepK{We retain one qualifying session per user.}
\apebKeepK{These criteria select active, exploration-heavy, short-horizon shopping sessions with observable query refinement: each retained session contains at least two queries and at least ten product or media interactions before a purchase within 30 minutes
of the initial search (Appendix~\ref{ap-benchmark-validity}).}

\subsection{History \& Query-Candidate Identification}\label{sec:bench-identify}
For users with retained sessions, we construct their heterogeneous histories and aligned query--candidate pairs.

\subsubsection{Heterogeneous History Curation}\label{sec:bench-history}
Heterogeneous long-term interactions provide rich signals for inferring user persona.
For a user \(u\) with historical interaction log $\mathcal{H}_u$ (Eq.~\ref{eq:history}),
we first filter out low--dwell-time records (Appendix \ref{ap-benchmark-history-dwell-filter}). 
We then consider two complementary types for the remaining interactions.

\textbf{(1) Product interactions.}
Each product interaction is defined as
\begin{equation}\label{eq:his-product-interact}
x_{u,\mathrm{product}}^k = (\mathrm{meta}_p^k,\ \mathrm{type}^k,\ \Delta t^k),
\end{equation}
where
\(
\mathrm{meta}_p^k = (\mathrm{id},\ \mathrm{cat},\ \mathrm{title},\ \mathrm{detail},\ \mathrm{features})
\)
encodes product metadata,   
\apebKeepK{\(\mathrm{type}^k\in\{\text{view},\text{like},\text{cart},\text{order}\}\), and \(\Delta t^k\) is dwell time.}

\textbf{(2) Media interactions.}
A media (video or livestream) interaction is represented as
\begin{equation}\label{eq:his-media-interact}
x_{u,\mathrm{media}}^k = (\mathrm{meta}_m^k,\ \Delta t^k),
\end{equation}
where
\(
\mathrm{meta}_m^k = (\mathrm{id},\ \mathrm{tags},\ \mathrm{title},\ \mathrm{detail},\ \mathcal{P}_{\mathrm{refer}}^k),
\)
and \(\mathcal{P}_{\mathrm{refer}}^k\) denotes the set of products referenced or mentioned in the media,
while \(\Delta t^k\) captures user engagement duration
\apebKeepC{(see Appendix~\ref{ap-benchmark-history-content})}.

\subsubsection{\smartqB \& \matchqB Query}\label{sec:bench-query}
We distinguish two query types from the query refinement trajectory based on their position and role.
For each user \(u\), let
\(
\mathcal{Q}_u^{\mathrm{search}} = (q_u^1, \ldots, q_u^{|\mathcal{Q}_u^{\mathrm{search}}|})
\)
denote the temporally ordered search queries in the retained search-to-order session, and let
\(p_u^{o}\) be the purchased product.
We define:

(i) \textbf{\smartqB query} \(q_u^{\mathrm{\smartq}}\): \apebKeepK{the earliest query in \(\mathcal{Q}_u^{\mathrm{search}}\) semantically aligned with \(p_u^{o}\),}
denoting an intentional but ambiguous goal (e.g. \emph{campus ootd}).

(ii) \textbf{\matchqB query} \(q_u^{\mathrm{\matchq}}\): the last query issued before purchase that is explicit in product type/attributes,
representing a well-specified intent (e.g. \emph{oversized long-sleeve}).

Query--product alignment is judged using GPT-4.1 with human verification. We discard cases containing only a single aligned query \apebKeepC{(Appendix~\ref{ap-benchmark-filter-session-distinct-query})}.
\apebKeepH{Blinded construction audits show strong agreement for both session alignment and intent/refined-query localization (Appendices~\ref{ap-benchmark-filter-session-rule3} and~\ref{ap-benchmark-filter-session-distinct-query}).}
Unlike \matchq, which primarily evaluates semantic retrieval~\cite{query-dataset-3-JDsearch},
\smartq reflects early-stage search behavior requiring intent inference~\cite{switch-query-3}.
\apebKeepD{The distinction is defined by the session trajectory, not a fixed semantic boundary---identical query forms can reflect different levels of ambiguity at different stages of refinement.}
Accordingly, \smartq queries stress an agent's capacity to resolve underspecified goals into actionable intent under personalization.

\subsubsection{Hard Candidate Identification}\label{sec:bench-candidate}
\apebKeepK{All unbought viewed products are candidates; at least nine share the purchase category.}
For the retained session of user \(u\), the hard candidate set is defined as
\begin{equation}\label{eq-hard-candidate}
\apebKeepK{\mathcal{C}_u^{\mathrm{hard}} =
(\mathcal{P}_u^{\mathrm{view}}\setminus\{p_u^{o}\}) \cup \{p_u^{o}\},}
\end{equation}
where each candidate \(c \in \mathcal{C}_u^{\mathrm{hard}}\) is represented by its product metadata
\(\mathrm{meta}_c\).
\apebKeepH{Candidate metadata comprises both commercial attributes, such as price and promotion, and fine-grained product attributes, including material and ingredients.}
\apebKeepH{Candidate-side ablations show that commercial attributes provide modest auxiliary value, whereas full metadata substantially outperforms commercial attributes alone (Appendix~\ref{ap-hard-candidate-controls}).}
\apebKeepH{This indicates that distinguishing \(p_u^{o}\) from \(\mathcal{C}_u^{\mathrm{hard}}\) 
requires agents to replicate the user's personalized,
preference-aligned comparison process over fine-grained attributes beyond commercial cues.}

\subsection{Benchmark Evaluation}\label{sec:bench-eval}
\benchs evaluates models under two query settings:

\textbf{Input.}
\apebKeepK{For query variant \(v\in\{\smartq,\matchq\}\):}
\begin{equation*}
\apebKeepK{\text{Input}_{u}^{v}
= (q_u^{v}, \mathcal{H}_u, \mathcal{C}_u^{\mathrm{hard}}).}
\end{equation*}

\textbf{Output and Metrics.}
\apebKeepK{We return five items and average the instance-level scores:}
\begin{equation*}
\apebKeepK{\begin{aligned}
(\hat c_{u,k})_{k=1}^{5}
&= \operatorname{Rank}_{\theta}(\text{Input}_{u}^{v})_{1:5},\\
\mathrm{Hit@K}(u)
&= \mathbb{I}[p_u^{o}\in\{\hat c_{u,k}\}_{k=1}^{K}],
\quad K\in\{1,5\}.
\end{aligned}}
\end{equation*}

\section{Benchmark Post-Processing}\label{sec:bench-post}
\textbf{Anonymization}
Before evaluation, user behavior data are anonymized to prevent personal information leakage. 
Raw user identifiers are removed, and each interaction sequence is mapped to a randomly assigned surrogate ID with no cross-split linkability. 
Temporal signals are discretized into coarse-grained time bins to avoid fine-grained behavioral tracing. 
We retain public item and content metadata to support heterogeneous personalization tasks.
Meanwhile, we apply automated personally identifiable information redaction to all free-form text---particularly OCR and ASR outputs---to remove explicit and implicit identifiers.
\apebKeepH{\benchs will be released through the project website\footnote{\url{https://voxxxx1874.github.io/APeB-Website/}} after compliance and privacy review, with data minimization; Appendix~\ref{ap-release-governance} details the release scope and governance.}

\textbf{Statistics} 
\apebKeepH{\benchf contains 5,648 retained English cases spanning major categories such as Personal Care, Clothing, Food \& Beverages, and Home Supplies.}
On average, each user has 187 historical interactions, including 82 product and 105 media interactions. These histories contain an average of 18 historical orders. 
\apebKeepK{After excluding \(>25\)-candidate sessions, candidate sets contain \(14.1\pm2.8\) items, including at least nine non-purchased products in the purchase category (Appendix~\ref{ap-hard-candidate-controls}).}

\section{Experiments and Analysis}
\begin{table*}[t]
\begin{center}
\begin{small}
\setlength{\tabcolsep}{3pt}
\renewcommand{\arraystretch}{0.96}
\resizebox{\textwidth}{!}{%
\begin{tabular}{lll@{\hspace{2mm}}cccccccc@{\hspace{2mm}}c}
\toprule
\textbf{\apebKeepB{Query}} & \textbf{Method} & \textbf{Metric (\%)}
& Gemini & 4.1mini & GPT4o & Qwen3 & R1 & GPT4.1 & GPT5.2 & GPT5.2Think & $\text{UniSAR}^*$ \\
\midrule
\multirow{8}{*}{\textbf{\apebKeepB{\smartqB}}}
& \multirow{2}{*}{\textbf{Single-Prompt}}
& Hit@1 & 22.1 & 23.6 & 24.9 & 22.7 & 24.2 & 23.9 & 25.6 & 24.2 & \textbf{25.9} \\
& & Hit@5 & 76.6 & 71.1 & 75.7 & 71.5 & 74.7 & 73.3 & \textbf{78.3} & 77.8 & 59.2 \\
\cmidrule(lr){2-12}
& \multirow{2}{*}{\textbf{ReAct}}
& Hit@1 & 24.2 & 26.0 & 26.3 & 24.6 & 24.8 & 25.8 & 26.2 & \textbf{26.4} & -- \\
& & Hit@5 & 76.2 & 72.2 & 75.8 & 71.1 & 73.9 & 77.9 & 77.5 & 77.3 & -- \\
\cmidrule(lr){2-12}
& \multirow{2}{*}{\textbf{Deerflow}}
& Hit@1 & 21.1 & 22.6 & 23.2 & 23.1 & 22.9 & 22.2 & 23.5 & 23.5 & -- \\
& & Hit@5 & 71.3 & 69.4 & 71.5 & 72.6 & 70.9 & 72.3 & 72.5 & 73.8 & -- \\
\cmidrule(lr){2-12}
& \multirow{2}{*}{\textbf{\ouragent}}
& Hit@1 & 25.2 & 23.9 & 26.3 & 25.0 & 26.2 & 24.9 & 26.4 & \textbf{26.9} & -- \\
& & Hit@5 & 76.9 & 72.2 & 75.7 & 75.0 & 76.3 & 74.8 & \textbf{78.5} & 78.2 & -- \\
\midrule
\multirow{4}{*}{\textbf{\apebKeepB{\matchqB}}}
& \multirow{2}{*}{\textbf{Single-Prompt}}
& Hit@1 & 33.9 & 34.4 & 37.0 & 35.8 & 36.2 & 34.5 & \textbf{37.4} & 37.0 & 29.2 \\
& & Hit@5 & 81.9 & 79.6 & 82.8 & 81.6 & 80.7 & 78.8 & \textbf{84.2} & 83.6 & 64.5 \\
\cmidrule(lr){2-12}
& \multirow{2}{*}{\textbf{ReAct}}
& Hit@1 & 35.7 & 36.4 & 37.8 & 37.3 & 36.0 & 40.4 & \textbf{43.3} & 42.8 & -- \\
& & Hit@5 & 81.8 & 79.7 & 84.2 & 82.7 & 81.7 & 83.2 & \textbf{84.4} & 83.4 & -- \\
\bottomrule
\end{tabular}
}
\end{small}
\end{center}
\caption{
\apebKeepA{Hit@1/5 performance for LLM workflows (Single-Prompt (60 category-prior selected records), ReAct, Deerflow and VQRA) and PPS model UniSAR under \smartq and \matchq queries.}
\apebKeepA{VQRA denotes the setting where the \smartq query has been rewritten with a shared backbone (GPT-4.1) using selected history context before evaluation.}}
\label{tab:experiment-overall-query-settings}
\end{table*}

We evaluate personalized shopping goal completion and intermediate reasoning on \benchf under \matchq and \smartq across \apebKeepE{8 LLMs} and their agentic variants, addressing the following questions:

\textbf{RQ1:} How do LLM workflows perform compared to recommender PPS methods?
We compare hit rates on \smartq and \matchq queries with different history/candidate sets. 

\textbf{RQ2:} Which types of flaws most adversely affect personalization tasks?  
We conduct a rubric analysis to attribute performance drop to intermediate \apebKeepA{reasoning} errors.

\textbf{RQ3:} How can the failures in RQ2 be addressed? 
\apebKeepA{We use a \ouragentf (\ouragentB) to demonstrate that explicit history utilization modules reduce this gap.}

\subsection{Experimental Setup}

\subsubsection{Evaluated Models}
We evaluate:
(1) PPS baseline: UniSAR \cite{pps-s-6}.
(2) LLMs: Qwen3-235B \cite{qwen3}, DeepSeek-R1 \cite{deepseekr1}; GPT-4/5 series \cite{gpt4, gpt5}, Gemini-2.5-Pro \cite{gemini2.5}.
(3) LLM Agents: ReAct \cite{ReAct}, Deerflow \apebKeepC{\cite{deerflow2025}}.
\apebKeepA{
    Appendix~\ref{ap-experiment-additional-baselines} discusses additional baselines.}

\subsubsection{Benchmark Utilization}
Due to the large semantic volume, we utilize \benchf :

\textbf{(1) UniSAR - Supervised Training} 
Following \citet{query-dataset-4-KuaiSAR}, we construct SAR training data from historical query--order pairs and isolated purchases. 
The training history is defined as \(\mathcal{H}_u^{\text{SAR}} = \{ i_u^k \}_{k=1}^{|\mathcal{H}_u^{\text{SAR}}|}\), where each interaction 
\(i_u^k \in \{ (q_u^k, p_u^k),\,p_u^k \}\).

\textbf{(2) LLM - Single Prompt} 
To fit histories, queries, and candidates within the LLM context window, we 
retain \(k\) records from \(\mathcal{H}_u\). 
\apebKeepA{We evaluate category-prior and time-prior history selection (same-category label associated with the intent query vs. most-recent records), keeping only essential fields per record (title, price).}
Details are in Appendix~\ref{ap-experiment-single-prompt}.

\textbf{(3) Agent - External Retrieval Database}
Agents use a retrieval tool backed by a database over history and candidate metadata.
At each step, agents issue queries to retrieve extra information beyond the prompt
\apebKeepC{(Appendix \ref{ap-experiment-agent}, \ref{ap-deerflow}).}

\subsubsection{Evaluation Method}\label{sec-exp-eval-output}

\begin{algorithm}[t]
\caption{Structured Product Recommendation Step}
\label{alg:structured-reco}

\textbf{Input:} \smartq/\matchq query $q$, histories $\mathcal{H}$, candidates $\mathcal{C}^{\mathrm{hard}}$

\textbf{Target inference:}
\apebKeepK{Infer user target \(\tau\) from \((q,\mathcal{H})\).}

\textbf{Relevant history selection:}
\apebKeepK{\(\mathcal{H}^\star \leftarrow\) histories salient to \(\tau\).}

\textbf{History summarization:}
\apebKeepK{\(\mathcal{F}=\{(f_i,\pi_i,\mathcal{H}_i^\star)\}_{i=1}^{m}\), where
\(f_i\) is a preference fact, \(\pi_i\) its relevance rank, and
\(\mathcal{H}_i^\star\subseteq\mathcal{H}^\star\) its supporting records.}

\textbf{Recommend items:}
\apebKeepK{\(\mathcal{R}^\star=((c_j,\rho_j,\mathcal{F}_j))_{j=1}^{5}\), where
\(j\) is the recommendation rank, \(c_j\) the candidate, \(\rho_j\) its rationale, and
\(\mathcal{F}_j\subseteq\mathcal{F}\) its supporting preferences.}

\end{algorithm}

\textbf{Structured Output:} To evaluate both task success and intermediate steps, Algorithm~\ref{alg:structured-reco} prompts models to produce structured outputs, including user target, history summarization, and recommend items.
For ReAct, reasoning steps $Re_i$ and retrieval queries $Rq_i$ at each step $i$ are logged.

\textbf{Outcome \& Reasoning:} Personalization performance is measured using Hit@1/5 on the recommended items $\mathcal{R}^\star$.
\apebKeepA{A model succeeds when it ranks the purchased item above user-considered alternatives.}
We further apply rubric evaluation to intermediate outputs using \emph{LLM-as-a-Judge}, scoring each component independently.
\apebKeepA{The intermediate rubrics are anchored to the same user session evidence: they diagnose whether failures arise from intent inference, history-grounded preference extraction, recommendation alignment, or multi-step reasoning.}
Detailed evaluation is in Section \ref{sec:exp-why-fail}.

\subsection{Overall Results and Analysis (RQ1)}

\subsubsection{LLMs Underperform on \smartqB Queries}\label{sec:exp-llm-fall}

\apebKeepA{Table~\ref{tab:experiment-overall-query-settings} shows that while single-prompt LLMs achieve strong Hit@5 across workflows under both query types, they slightly underperform UniSAR on Hit@1 for \smartq queries.}
This highlights their \textbf{weakness in precise personalized ranking under vanilla goals and noisy histories}, where user modeling outweighs semantic matching.
\apebKeepH{A scaling experiment on Qwen series further shows that model scaling improves Hit@1 but does not eliminate the persistent intent--refined gap (Appendix~\ref{ap-experiment-qwen-scale}).}

We further evaluate agents with multi-step reasoning and tool use.
\apebKeepA{In Table~\ref{tab:experiment-overall-query-settings}, the \matchq-query block shows that ReAct agents \apebKeepJ{generally} outperform single-prompt LLMs, while the \smartq-query block shows that ReAct yields only marginal gains and Deerflow shows little benefit or even degradation.}
These results indicate that \textbf{increased semantic capacity alone does not guarantee improved personalization}: when user goals are weakly specified, additional planning and retrieval steps can instead amplify reasoning errors. \apebKeepC{Detailed analysis is in Section~\ref{sec:exp-why-fail}}.
\apebKeepK{Additional paired bootstrap analysis with confidence intervals is provided in Appendix~\ref{ap-statistical-reliability}.}

\subsubsection{Hard Candidates and Noisy History}

\textbf{\apebKeepA{LLMs struggle to benefit from history.}}
Table \ref{ablation-history-length} shows that with no-history, models demonstrate basic semantic matching ability, while \smartq remains more challenging.
When user history is introduced, trained UniSAR achieves substantial gains.
\apebKeepA{Short and long category-prior histories (@6/@60 CATE) yield very similar Hit@1 under \smartq queries;
while the refined-query block shows larger improvements once intent ambiguity is reduced.}
\apebKeepA{Comparisons between \smartq and \matchq queries imply that history possesses usable preference signal but current LLM workflows do not naturally benefit 
possibly because LLMs do not extract recoverable preferences under \smartq queries.}
\apebKeepH{History interventions show that user-matched histories outperform shuffled and counterfactual histories, while simply expanding the context yields limited benefit (Appendices~\ref{ap-experiment-history-stress} and~\ref{ap-experiment-contamination-counterfactual}).}
\apebKeepH{Replacing raw histories with explicitly extracted preferences yields only modest gains for GPT-4.1-mini and Qwen3 and leaves GPT-5.2 unchanged, showing that a separate preference-extraction step alone provides limited end-to-end improvement. The remaining difficulty may stem from incomplete recovery of task-relevant preferences, ineffective use of those preferences in fine-grained candidate comparison (Table~\ref{ap-tab:explicit-preference-representation}).}
Further analysis in Section \ref{sec:exp-rq3-vqra} provides stepwise evidence.

\begin{table}[t]
\centering
\scriptsize
\setlength{\tabcolsep}{3pt}
\renewcommand{\arraystretch}{0.95}
\resizebox{\columnwidth}{!}{%
\begin{tabular}{lcccc}
\toprule
\textbf{\apebKeepB{History}} & UniSAR & Qwen3 & GPT5.2 & R1 \\
\midrule
\multicolumn{5}{l}{\textit{\apebKeepB{\smartqB query}}} \\
\apebKeepB{@0}        & 14.1 & 21.2 & 23.8 & 22.2 \\
\apebKeepB{@6 Cate}   & --   & 21.6 & 24.7 & 23.7 \\
\apebKeepB{@60 Cate}
          & $\uparrow$\,25.9 {\scriptsize(+84\%)}
          & $\uparrow$\,22.7 {\scriptsize(+7.1\%)}
          & $\uparrow$\,25.6 {\scriptsize(+7.6\%)}
          & $\uparrow$\,24.2 {\scriptsize(+9.0\%)} \\
\midrule
\multicolumn{5}{l}{\textit{\apebKeepB{\matchqB query}}} \\
\apebKeepB{@0}        & 18.3 & 30.4 & 33.4 & 31.9 \\
\apebKeepB{@6 Cate}   & --   & 33.1 & 35.2 & 35.0 \\
\apebKeepB{@60 Cate}
          & $\uparrow$\,29.2 {\scriptsize(+59.6\%)}
          & $\uparrow$\,35.8 {\scriptsize(+17.8\%)}
          & $\uparrow$\,37.4 {\scriptsize(+12.0\%)}
          & $\uparrow$\,37.0 {\scriptsize(+16.0\%)} \\
\bottomrule
\end{tabular}
}
\caption{
\apebKeepA{History ablation: Hit@1 (\%) on \smartq/\matchq queries. @0 = no history; @k Cate = k category-prior records; arrows = change vs.\ @0.}}
\label{ablation-history-length}
\end{table}

\begin{table}[t]
\centering
\scriptsize
\renewcommand{\arraystretch}{1.0}
\begin{tabular}{lccccc}
\toprule
\textbf{\apebKeepB{Candidates}} & UniSAR & Qwen3 & GPT4.1 & GPT5.2 & R1 \\
\midrule
\apebKeepB{Random} & 81.3 & 84.8 & 86.3 & \textbf{95.9} & 93.1 \\
\apebKeepB{Hard}   & \textbf{25.9} & 22.7 & 23.9 & 25.6 & 24.2 \\
\bottomrule
\end{tabular}
\caption{
\apebKeepA{Candidate ablation on \smartq queries: Hit@1 (\%) with random vs.\ hard candidates.}}
\label{tab:ablation-candidates}
\end{table}

\textbf{\apebKeepA{Personalization choice lies beyond semantic matching.}}
Table \ref{tab:ablation-candidates} shows that, even under vague queries, LLMs outperform the UniSAR baseline when evaluated with random candidate sets, which are sampled from all possible candidates in the dataset. This gain mainly stems from LLMs' ability to exploit coarse semantic cues in underspecified intents to filter out obviously mismatched negatives and locate the 
\apebKeepA{query-aligned purchased} item. State-of-the-art models (e.g., GPT-5.2) further widen this margin over older or open-source models (Qwen3, GPT-4.1). 
Under hard candidates, both gains over UniSAR and inter-LLM performance gaps largely disappear, indicating that \apebKeepD{current LLMs struggle to discriminate fine-grained attributes among user-viewed candidates.}
\apebKeepA{
This also shows that: random negatives often require coarse semantic matching rather than true personalization. Hard candidates drawn from the same shopping trajectory better test whether a model can distinguish the purchased item based on user-specific preferences.
}

\subsection{Rubrics on Intermediate Reasoning (RQ2)}\label{sec:exp-why-fail}

\begin{figure*}[t]
\centering

\begin{subfigure}[t]{0.32\textwidth}
  \centering
  \includegraphics[width=\linewidth]{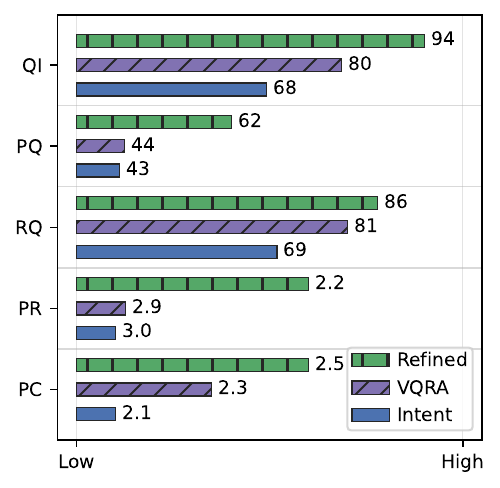}
\caption{Averaged scores over LLM models \protect\newline on \smartqB/\apebKeepA{\ouragent}/\matchqB query}\label{fig:ep-rubric-match-smart-llm}
\end{subfigure}
\begin{subfigure}[t]{0.32\textwidth}
  \centering
  \includegraphics[width=\linewidth]{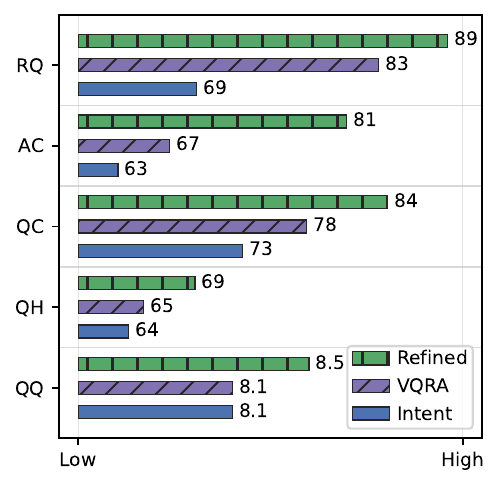}
        \caption{Averaged scores over ReAct models on \smartqB/\apebKeepA{\ouragent}/\matchqB query}\label{fig:ep-rubric-match-smart-agent}
\end{subfigure}
\begin{subfigure}[t]{0.32\textwidth}
  \centering
  \includegraphics[width=\linewidth]{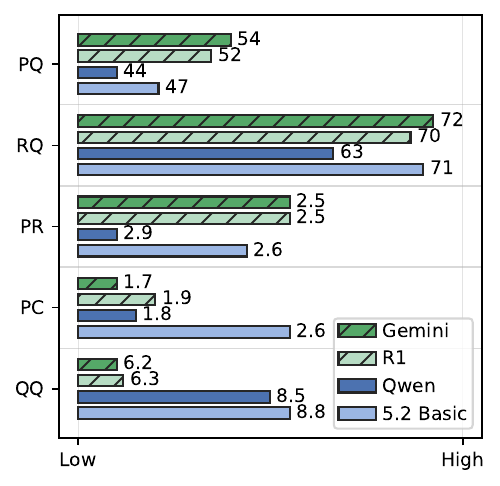}
    \caption{Per-model analysis of intermediate \protect\newline reasoning components on \smartqB query}
    \label{fig:ep-rubric-reasoning-model}
\end{subfigure}
\caption{Rubric scores for evaluating models across intermediate stepwise outputs
(\text{QI}: Query Inference; \text{PQ}: Preference Quality; \text{RQ}: Recommendation Quality;
\apebKeepA{\text{PR}: Minimum Observed-Purchase Preference Rank} $\downarrow$; \text{PC}: Total Observed-Purchase Preference Count; 
\text{AC}: Agent Coherence; \text{QC}: Query--Candidate; \text{QH}: Query--History; \text{QQ}: Query Quantity).
\apebKeepB{Diagnostic scores are averaged over Gemini-2.5-Pro, GPT-5.2, and GPT-4.1 judges.}}
    \label{fig:exp-rubric}
\end{figure*}

We conduct rubric-based evaluation of intermediate outputs in Section~\ref{sec-exp-eval-output} to diagnose why 
\apebKeepA{LLMs and ReAct agents underperform on \smartq queries} and \apebKeepD{to compare behaviors across LLMs.}
\apebKeepA{We first infer a diagnostic goal proxy $g$ from in-session interactions (Eq.\ref{eq:session}) (Appendix \ref{ap-experiment-rubric-goal}), and evaluate each output against $g$ using \emph{LLM-as-a-Judge}.}
\apebKeepA{The diagnostic scores are averaged across Gemini-2.5-Pro, GPT-5.2, and GPT-4.1 judges with human verification.}
\apebKeepH{Appendix~\ref{ap-experiment-judge-robustness} maps the four stages below to their evaluated outputs and diagnostic criteria.}

\textbf{Multi-Level Rubric}
\apebKeepD{We decompose the evaluation into stepwise complementary aspects (Appendix~\ref{ap-experiment-rubric-aspect}):}
\apebKeepH{(1) \emph{Query Inference (QI)}---whether the model understands the query and identifies its intended candidate category;}
\apebKeepH{(2) \emph{Historical Preference Reasoning}---via i. \emph{Preference Quality (PQ)}, assessing whether extracted preferences $\mathcal{F}$ faithfully summarize the query-specific aspects the user cares about from history, and ii. Salience Statistics: minimum rank (\emph{PR}) and total count (\emph{PC}) of preferences supporting the purchased item;}
\apebKeepH{(3) \emph{Recommendation Quality (RQ)}---whether the recommended items $\mathcal{R}^\star$ and their rationales $\rho$ address the query and align with the user's preferences;}
(4) \emph{Agent Reasoning (AR)}---coherence of reasoning steps $Re_i$ (AC) and retrieval utility, including candidate/history queries (QC, QH) and query quantity (QQ), toward achieving $g$ without hallucination.
\apebKeepA{Rubric--Hit@K alignment and judge-sensitivity checks in Appendix \ref{ap-experiment-judge-robustness} and \ref{ap-experiment-rubric-verification} support using these scores as diagnostic evidence.}

\textbf{Intermediate Results}
Figure~\ref{fig:exp-rubric} shows rubric scores for \smartq and \matchq queries under single-prompt and ReAct workflows. We observe:

\textbf{\emph{(1) Failures originate from intent understanding.}} 
As shown in Figure~\ref{fig:ep-rubric-match-smart-llm}, \emph{query inference} performance drops substantially on \smartq compared to \matchq queries, undermining downstream preference grounding and leading to degraded \emph{recommendation quality}. 
\apebKeepK{Notably, \emph{preference reasoning} remains limited even under explicit \matchq queries, indicating that identifying and summarizing relevant signals from heterogeneous histories remain an open challenge, even with clarified goals.}
Under vague intents, inferring meaningful historical preferences is particularly difficult.
 
\textbf{\emph{(2) Agentic workflows amplify strengths and weaknesses.}} 
Comparing Figures \ref{fig:ep-rubric-match-smart-llm} and \ref{fig:ep-rubric-match-smart-agent}, multi-step retrieval improves recommendation quality under explicit \matchq queries but offers little benefit for \smartq queries. Under vague intents, low reasoning coherence yields ineffective retrieval queries, causing agentic reasoning to compound intent errors, consistent with \apebKeepA{the \smartq-query block of Table~\ref{tab:experiment-overall-query-settings}}. Moreover, retrieving relevant signals from user histories is substantially harder than retrieving candidate information under both query types.
\apebKeepK{Appendix~\ref{ap-react-failure-analysis} further analyzes ReAct's retrieval utility, round depth and reasoning coherence.}

\textbf{\emph{(3) Models exhibit different intermediate behaviors.}}
\apebKeepD{Figure~\ref{fig:ep-rubric-reasoning-model} shows that models with comparable Hit@1 diverge in their intermediate reasoning steps.}
Advanced reasoning models (e.g., Gemini-2.5-Pro, DeepSeek-R1) achieve better intermediate performance in history preference reasoning and recommendation quality. 
\apebKeepK{They rank purchase-supporting preferences earlier (lower PR) but find fewer (lower PC).}
They also issue fewer retrieval queries under ReAct (lower QQ).
\apebKeepD{These might become detrimental especially under \smartq queries where insufficient preference coverage undermines recommendation.}

\subsection{\ouragentB: Refine \smartqB Query First (RQ3)}\label{sec:exp-rq3-vqra}
Section~\ref{sec:exp-why-fail} and Figure~\ref{fig:exp-rubric} attribute the performance drop on \smartq queries to weak query inference due to underexplored histories. 
\apebKeepA{To isolate this gap, we evaluate a history-conditioned query-refinement agent, \ouragent, in which a shared backbone (GPT-4.1) first rewrites the \smartq query using selected same-category histories and the LLM backbone is then evaluated with the rewritten query under the single-prompt pipeline.
}%
\apebKeepA{Table~\ref{tab:experiment-overall-query-settings} shows that \ouragent improves Hit@1, with the largest gain on GPT-5.2Think.}
\apebKeepA{Appendix~\ref{ap-experiment-VQRA} shows similar results with the same-backbone-rewrite variant.}
\apebKeepA{Rubric scores in Figure~\ref{fig:ep-rubric-match-smart-llm}, \ref{fig:ep-rubric-match-smart-agent} suggest that query refinement improves \emph{query inference (QI)} toward the shopping intent, leading to gains in \emph{recommendation quality (RQ)}.}
\apebKeepA{These indicate that inferring intent from vague queries and noisy histories remains underutilized: 
with explicit history grounding, LLMs already produce useful refinements, suggesting headroom for personalized agent design.
}

\section{Conclusion}
We introduce \benchs, a benchmark for agentic personalization that evaluates personalized product search under underspecified queries and noisy interaction histories. Through \benchs, we show that current LLMs handle explicit queries well but struggle in early-stage personalization, where they must infer intent, recover preferences, and choose among competing alternatives. 
\apebKeepJ{Rubric analysis links this gap to ineffective history use: models struggle to extract task-relevant, well-grounded preferences from user histories and apply them to candidate selection.}
Furthermore, \ouragent, a simple history-aware query-refinement pipeline, improves performance across backbones, highlighting the value of explicit history-utilization modules. \benchs serves as a testbed for robust user modeling in real-world personalization.

\section*{Limitations}
\apebKeepF{\benchs is drawn from a single content-commerce platform; behavior patterns, query styles, and candidate structures may not transfer to other e-commerce ecosystems or non-shopping personalization domains.
All evaluation is static and offline: fixed candidate sets and recorded purchases do not reflect real-time inventory, pricing, promotions, or multi-turn interaction.
Observed purchases serve as ground-truth preference proxies but are confounded by ranking exposure, stock availability, and platform mediation.
We explore category- and time-based history selection without exhaustively studying advanced memory architectures or adaptive context-length strategies.
Rubric analysis relies on LLM-as-a-Judge, which may introduce systematic biases despite strong human agreement.
The benchmark covers limited locale and language; intent expression and attribute salience may vary across cultures.}
\apebKeepH{The retained benchmark focuses on active, exploration-heavy shopping sessions with an attributable query--purchase outcome within a 30-minute window; future work may extend the attribution protocol to sparse and cross-day journeys.}
\apebKeepH{The evaluation input represents heterogeneous media histories through postprocessed summaries, captions, anonymized OCR/ASR text, and referenced products; future work may add raw-media inputs.}

\section*{Ethical Considerations}
\apebKeepA{This work studies personalization benchmarks derived from postprocessed behavioral shopping records. Its positive impact is to make evaluation of preference-conditioned product search more realistic and auditable, especially when users issue ambiguous queries. The main risks are privacy leakage from long histories or product/media text, re-identification through rare behavior patterns, category or platform bias, and misuse of preference inference for manipulative over-personalization.}
\apebKeepH{We mitigate these risks through a layered release pipeline that exposes only postprocessed, anonymized benchmark instances rather than raw logs. We apply coordinated semantics-preserving substitution across queries, histories, and candidate descriptions: source-specific expressions are replaced with semantically matched text, while query--history--candidate alignment and task-critical product attributes---including product type, size, material, and ingredients---are retained. We further remove direct identifiers and platform-private fields, assign surrogate identifiers without cross-split linkability, coarsen temporal and price signals, minimize retained interactions, and redact personally identifiable content from free-form OCR/ASR text.}

\section*{Acknowledgments}
We thank the reviewers for their constructive comments and suggestions.
\apebKeepH{Garry Yang}, Zizhe Chen, Deyu Zou, Yongqiang Chen, and James Cheng were supported by a CRF (No. C2005-24Y) from the RGC of Hong Kong.
This work was a collaboration between Husky Data Lab at CUHK and ByteDance, supported by a ByteDance University Collaboration Project Grant.
Generative AI assistants were used for language polishing, writing and coding assistance, and preparation of submission materials. All AI-assisted content and outputs were reviewed and verified by the authors, who take full responsibility for the work.

\bibliography{apeb_emnlp_2026}

\appendix
\let\prompt\relax
\let\endprompt\relax
\newcommand{\promptsep}{\par\noindent\rule{\linewidth}{0.4pt}\par}
\newenvironment{prompt}{%
  \begin{quote}
  \small\ttfamily\raggedright
  \setlength{\parindent}{0pt}
  \setlength{\parskip}{0.25\baselineskip}
}{%
  \end{quote}
}

\section{Benchmark Collection}\label{ap-benchmark-collect}
\subsection{Holistic User Action Platform}\label{ap-benchmark-platform}
\begin{figure}[t]
    \centering
    \includegraphics[height=6.1cm]{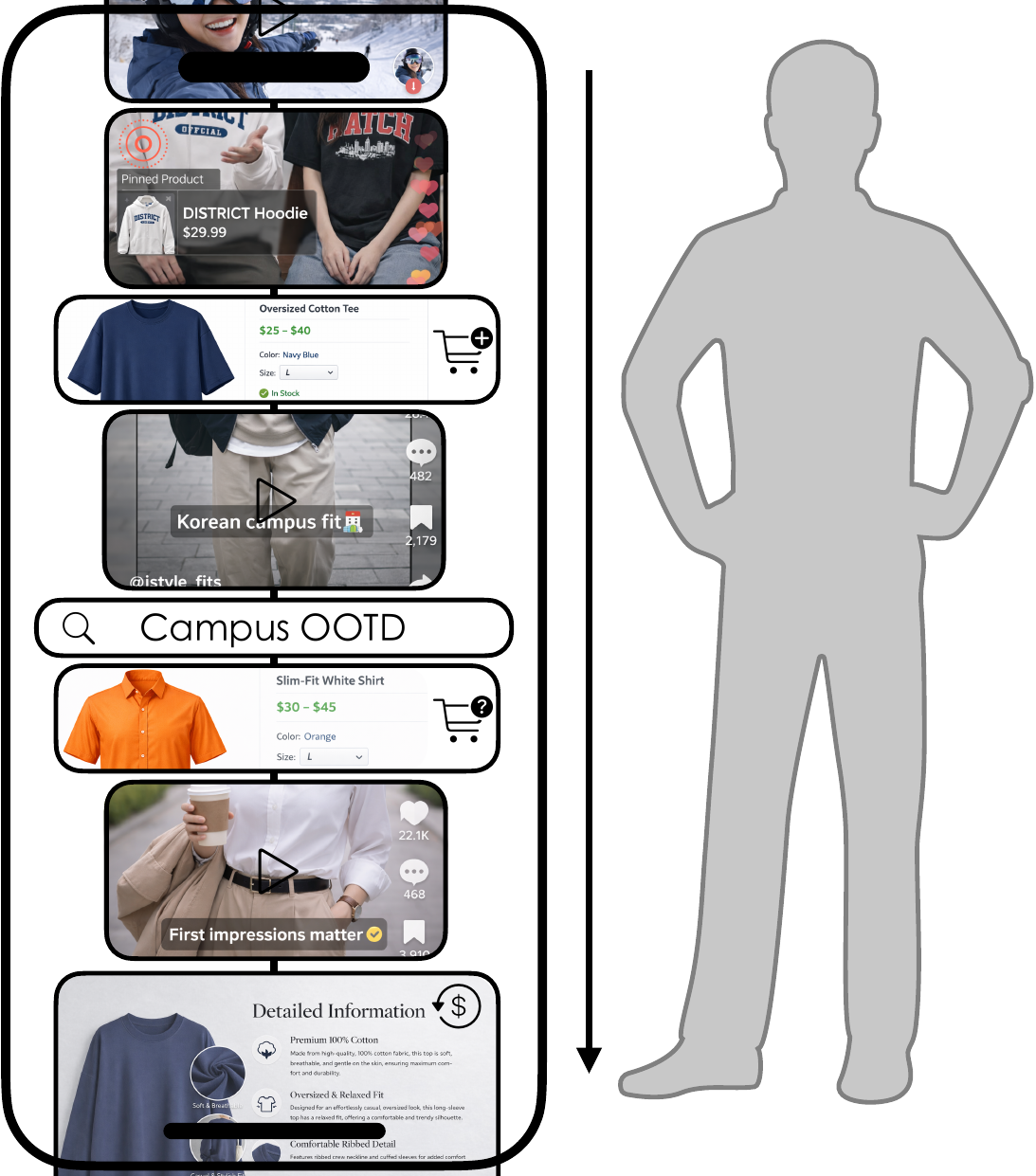}
    \caption{
    User interaction timeline on the base platform.
    }
    \label{fig:ap-benchf-timeline}
\end{figure}

The base platform underlying \benchf hosts billions of short-form videos and livestreams and serves as a major venue for commerce. Users primarily consume video and livestream content via continuous scrolling, but can seamlessly transition to search for videos, livestreams, or products, as well as navigate directly to product detail pages embedded within promotional content. The queries collected in \benchf are restricted to those exhibiting at least weak e-commerce intent. We further illustrate this user interaction timeline in Figure~\ref{fig:ap-benchf-timeline}. All user data in \benchf are collected in compliance with the platform's privacy policies and contractual data-use requirements.

\subsection{\apebKeepH{Daily Active-User Filtering and Session Sampling}}\label{ap-benchmark-filter-user}
\apebKeepH{We process each day in the 30-day session-sampling window separately. For each sampling day, we retain users who have at least five product or media interactions on every day of the preceding 30 days. We then identify qualifying search-to-order sessions occurring on that sampling day within this daily active-user pool (Section~\ref{sec:benchmark-discover}). Across the 30 sampling days, if a user has multiple qualifying sessions, we retain the earliest one. For each retained session, we collect up to 60 days of historical interactions preceding the timestamp of its first search query, including e-commerce-related search queries, livestreams, videos, and product interactions.}

\subsection{Session Filtering}\label{ap-benchmark-filter-session}
We detail the implementation of session filtering rules introduced in Section~\ref{sec:bench-filter}.
\subsubsection{Rule-based Session Filtering}\label{ap-benchmark-filter-session-only-rule}
We first apply deterministic, rule-based filtering using SQL over raw interaction logs. 
\apebKeepK{SQL retains sessions satisfying: (i) at least ten product or media interactions (Eq.~\ref{eq:extensive-browsing}); (ii) at least two distinct queries (Eq.~\ref{eq:query-refine}); and (iii) at least nine non-purchased viewed products sharing the purchased item's category. This category count is a session-level floor; after a session passes, its other viewed products remain in the final candidate set.}
These criteria remove shallow or accidental purchases and ensure that retained sessions reflect non-trivial exploration and intent evolution.
\subsubsection{\apebKeepK{Intent--purchase alignment verification}}\label{ap-benchmark-filter-session-rule3}
\apebKeepK{For sessions passing the SQL filters, GPT-4.1 verifies that the purchased item is semantically consistent with at least one issued query. Human review validates this query--purchase decision.}
\apebKeepH{We use a shared audit sample of 100 de-identified sessions stratified by product category and query-sequence length. Two blinded annotators inspect the query trajectory, viewed-item metadata, and purchased-item metadata; a third annotator resolves disagreements.}
\apebKeepK{For query--purchase alignment, GPT-4.1 agrees with the adjudicated human decision on 97\% of sessions, raw human--human agreement is 96\%, and GPT-5.2 agreement is 97\% under the same inputs.}
Sessions failing this semantic consistency check are discarded; if multiple purchased items satisfy the criterion, we retain only one to ensure a unique target per session.

\subsubsection{Distinct \smartq and \matchq}\label{ap-benchmark-filter-session-distinct-query}
Given a session $u$ with query sequence \apebKeepK{$\mathcal{Q}_u^{\mathrm{search}}=\{q_u^1,\dots,q_u^{T_u}\}$} and purchased product $p_u^{o}$, we locate two aligned queries using GPT-4.1 by prompting it to \emph{return indices} in \apebKeepK{$\mathcal{Q}_u^{\mathrm{search}}$}:
\[
\begin{aligned}
t_{\smartq}=\min\{t:\ &q_u^t \text{ is semantically aligned}\\
&\text{with } p_u^{o}\},
\end{aligned}
\]
the earliest aligned query, i.e., \smartq, and
\[
\begin{aligned}
t_{\matchq}=\max\{t:\ &q_u^t \text{ is aligned and explicit}\\
&\text{in product type/attributes}\},
\end{aligned}
\]
the last explicit aligned query before purchase, i.e., \matchq.
We then set $q_u^{\smartq}=q_u^{t_{\smartq}}$ and $q_u^{\matchq}=q_u^{t_{\matchq}}$.

To ensure \emph{distinct} intents, we apply a minimal rule-based check after GPT-4.1 localization:
we discard session $u$ if $q_u^{\smartq}$ and $q_u^{\matchq}$ are identical after normalization (lowercasing and whitespace/punctuation stripping), i.e.,
$\mathrm{norm}(q_u^{\smartq})=\mathrm{norm}(q_u^{\matchq})$.
This removes cases where the session contains only a single aligned query (or duplicated re-issues) and thus lacks an observable intent refinement trajectory.

\apebKeepH{For the same 100 stratified sessions, annotators identify the earliest aligned ambiguous query and the latest aligned explicit query from the de-identified query sequence, relevant product metadata, and purchased item.}
\apebKeepH{GPT-4.1 agrees with the adjudicated exact index pair on 93\% of sessions, raw human--human agreement is 91\%, and GPT-5.2 exact-pair agreement is 93\%.}

\subsection{Heterogeneous History Curation}\label{ap-benchmark-history}
We curate heterogeneous user histories by unifying product-centric and media-centric interaction records into a structured sequence, enabling fine-grained modeling of user intent and preference evolution.

\subsubsection{Dwell-time filtering}\label{ap-benchmark-history-dwell-filter}
We discard video and product view interactions with dwell time shorter than 10 seconds, retaining only interactions that reflect meaningful user engagement. 

\subsubsection{Heterogeneous history content}\label{ap-benchmark-history-content}
\paragraph{Product interactions.}
Each product interaction \(x_{u,\mathrm{product}}^k\) (Eq.~\ref{eq:his-product-interact}) corresponds to an explicit user action on an item.
The product metadata \(\mathrm{meta}_p^k\) consist of an \emph{encrypted} product identifier \(\mathrm{id}\), a categorical taxonomy label \(\mathrm{category}\), and textual fields \(\mathrm{title}\) and \(\mathrm{detail}\), where \(\mathrm{detail}\) is parsed from the product's structured HTML page.
The \(\mathrm{feature}\) field contains a set of salient product attributes extracted by platform-level product understanding pipelines.
The interaction type \(\mathrm{type}^k \in \{\text{view}, \text{like}, \text{cart}, \text{order}\}\) reflects increasing engagement intensity, and \(\Delta t^k\) denotes the dwell time associated with the interaction event.

\paragraph{Media interactions.}
Each media interaction \(x_{u,\mathrm{media}}^k\) (Eq.~\ref{eq:his-media-interact}) represents user engagement with a short video or livestream.
The media metadata \(\mathrm{meta}_m^k\) include an \emph{encrypted} media identifier \(\mathrm{id}\), content \(\mathrm{tags}\), and a \(\mathrm{title}\).
The \(\mathrm{detail}\) field is constructed from platform-generated multimodal summaries, including captions and anonymized OCR/ASR transcripts, to capture the semantic content of the media.
The referenced product set \(\mathcal{P}_{\mathrm{refer}}^k\) consists of products explicitly linked or automatically annotated as mentioned within the media content.
The engagement duration \(\Delta t^k\) measures the user's viewing or participation time.
\apebKeepH{The contents of interacted media are represented by textual surrogates, such as summaries, captions, anonymized OCR/ASR text, and referenced products.}
\apebKeepH{The media summaries retain semantic and stylistic cues that can support personalized candidate selection. A lexical audit finds that 98.2\% of audited media descriptions contain at least one camera/editing, temporal, presenter-style, or atmosphere marker. Moreover, media-summary histories achieve Hit@1 of 24.1 versus 24.2 with full histories for R1 and 22.3 versus 22.7 for Qwen3 (Table~\ref{ap-tab:history-stress-tests}), indicating that the summaries preserve sufficient user-specific evidence for personalization in this setting.}

\subsubsection{\texorpdfstring{\apebKeepA{Release Scope and Governance}}{Release Scope and Governance}}
\label{ap-release-governance}
\apebKeepH{\benchf will be released through the \href{https://voxxxx1874.github.io/APeB-Website/}{project website} as an anonymized benchmark package after privacy and compliance review. The package is designed for research evaluation and excludes fields that create avoidable re-identification risk. Table~\ref{ap-tab:release-field-scope} summarizes the planned release boundary.}

\begin{table}[H]
\centering
\small
\begin{tabularx}{\linewidth}{@{}>{\raggedright\arraybackslash}p{0.29\linewidth}>{\raggedright\arraybackslash}X@{}}
\toprule
\textbf{\apebKeepH{Category}} & \textbf{\apebKeepH{Content and handling}} \\
\midrule
\apebKeepH{Task records} & \apebKeepH{Postprocessed queries, candidate sets, task labels, and purchased-item labels.} \\
\midrule
\apebKeepH{Preference context} & \apebKeepH{Product and media metadata, summaries, referenced products, and key attributes.} \\
\midrule
\apebKeepH{Linkable fields} & \apebKeepH{Entity expressions are substituted consistently across linked fields; identifiers are replaced with surrogates and timestamps are coarsened.} \\
\midrule
\apebKeepH{Raw/private data} & \apebKeepH{Raw logs, media, and unredacted text; direct identifiers; device/location/profile data; and platform internals.} \\
\bottomrule
\end{tabularx}
\caption{\apebKeepH{Planned website release boundary for \benchf. The package will include postprocessed task records and preference context, transform linkable fields, and exclude raw or private data.}}
\label{ap-tab:release-field-scope}
\end{table}

\paragraph{\apebKeepH{Semantic-preserving entity substitution.}}
\apebKeepH{Brand names, product names, and related source-specific expressions are replaced with semantically matched substitutes. A shared within-instance mapping is applied consistently across queries, histories, and candidate descriptions, retaining their relations and task-critical attributes such as product type, size, material, and ingredients.}

\paragraph{\apebKeepH{Privacy and linkage checks.}}
\apebKeepH{Before the website release, each benchmark instance will be screened for raw user/account identifiers, cross-split persistent user identifiers, fine-grained timestamps, raw event-log payloads, and device, IP, GPS, or account-profile attributes.}

\paragraph{\apebKeepH{Record and text checks.}}
\apebKeepH{Exact duplicate product exposures are collapsed where needed for candidate construction. Free-form product and media text is scanned for phone numbers, emails, addresses, usernames, URLs, and other unique identifiers; sampled records containing OCR/ASR-derived text receive additional spot checks.}
\paragraph{\apebKeepH{Data minimization.}}
\apebKeepA{Data minimization is applied at both the user and field levels. We retain only the history window and shopping-session evidence needed to define the benchmark task, cap the number of retained interactions, and omit raw events outside the benchmark construction window. In the retained benchmark, users average 187 historical interactions; under the source-log accounting used for governance, the retained records correspond to about 200 interactions per user per month, approximately 10\% of shopping-related activity.}

\paragraph{\apebKeepH{Intended use and restrictions.}}
\apebKeepA{The benchmark is intended for research evaluation of personalized product search, preference reasoning, history-aware ranking, and agentic shopping workflows. It is not intended for identifying individuals, reconstructing behavioral timelines, building user profiles outside the benchmark task, joining records with external datasets, production ad targeting, or optimizing manipulative personalization strategies. Dataset users should agree not to attempt re-identification, linkage attacks, demographic or sensitive-attribute inference about individuals, or redistribution outside the approved access terms.}

\subsection{\apebKeepH{Benchmark Validation and Scope}}\label{ap-benchmark-stat}
\apebKeepH{We characterize the retained benchmark records, validate hard-candidate construction and purchase recoverability, analyze activity and temporal properties, and illustrate the shared task structure across personalization settings.}
\benchs covers \apebKeepB{5,648} cases with the category distribution shown in Table \ref{ap-tab:category-distribution}\apebKeepB{.}
\apebKeepB{As Section~\ref{sec:bench-post} shows, retained users average 187 historical interactions, including 82 product interactions and 105 media interactions, with 18 historical orders on average; candidate sets contain 14.1 items on average.}

\begin{table}[t]
\centering
\small
\begin{tabular}{l c}
\toprule
\textbf{Category} & \textbf{Percentage (\%)} \\
\midrule
Personal Care      & 28.0 \\
Clothing           & 20.2 \\
Food \& Beverages            & 15.2 \\
Home Supplies      & 11.8 \\
Consumer Electronics         & 8.9 \\
Sports \& Outdoor            & 5.1 \\
Baby \& Maternity Products   & 3.7 \\
Others                & 7.1 \\
\bottomrule
\end{tabular}
\caption{Category distribution of purchased products in the benchmark\apebKeepB{; lower-frequency categories are grouped under Others}.}
\label{ap-tab:category-distribution}
\end{table}

\subsubsection{\texorpdfstring{\apebKeepH{Hard-Candidate Controls and Purchase Recoverability}}{Hard-Candidate Controls and Evidence for Purchase Recoverability}}
\label{ap-hard-candidate-controls}
\apebKeepK{We first verify that hard-candidate sets are valid, size-controlled, de-duplicated, and drawn from sessions satisfying the same-category floor (Table~\ref{ap-tab:hard-candidate-controls}). We then audit whether user histories provide evidence for the observed purchase and whether candidate decisions use metadata beyond commercial cues, including fine-grained attributes such as size, material, and ingredients. Human evaluation calibrates recoverability among these viewed alternatives.}

\begin{table}[H]
\centering
\footnotesize
\begin{tabularx}{\linewidth}{@{}>{\raggedright\arraybackslash}p{0.22\linewidth}>{\raggedright\arraybackslash}X@{}}
\toprule
\textbf{\apebKeepH{Control}} & \textbf{\apebKeepH{Retained setting}} \\
\midrule
\apebKeepK{\textbf{Set size}} & \apebKeepK{\(14.1 \pm 2.8\) items; discard sessions with \(>25\) candidates.} \\
\midrule
\apebKeepK{\textbf{Category floor}} & \apebKeepK{\(\geq 9\) non-purchased viewed products in the purchase category.} \\
\midrule
\apebKeepH{\textbf{Validity}} & \apebKeepH{\(<1\%\) invalid, unavailable, or incomplete items.} \\
\midrule
\apebKeepH{\textbf{Deduplication}} & \apebKeepH{Collapse duplicate encrypted product IDs.} \\
\midrule
\apebKeepH{\textbf{Target label}} & \apebKeepH{One purchased-item label per session.} \\
\bottomrule
\end{tabularx}
\caption{\apebKeepH{Hard-candidate construction controls in \benchf}}
\label{ap-tab:hard-candidate-controls}
\end{table}

\paragraph{\apebKeepH{Purchase-target validity audit.}}
\apebKeepH{Tables~\ref{ap-tab:purchase-hard-negative-distinguishability} and~\ref{ap-tab:preference-extraction-validity} test whether historical preferences distinguish the purchase from viewed hard negatives through history augmentation and explicit preference extraction. The results favor history-based evidence, with the gain preserved when candidate price is removed, showing that prior interactions contain evidence for the observed choice.}

\begin{table}[t]
\centering
\small
\setlength{\tabcolsep}{3pt}
\begin{tabularx}{\linewidth}{@{}>{\raggedright\arraybackslash}Xcc@{}}
\toprule
\textbf{\apebKeepH{Input evidence}} & \textbf{\apebKeepH{$>60\%$}} & \textbf{\apebKeepH{$>80\%$}} \\
\midrule
\apebKeepH{Query only} & \apebKeepH{79} & \apebKeepH{51} \\
\apebKeepH{Query + history} & \apebKeepH{93} & \apebKeepH{63} \\
\apebKeepH{Query + history, no price} & \apebKeepH{91} & \apebKeepH{59} \\
\bottomrule
\end{tabularx}
\caption{\apebKeepH{Percentage of sessions above each pairwise purchase win-rate threshold ($N=500$); ``no price'' removes candidate price.}}
\label{ap-tab:purchase-hard-negative-distinguishability}
\end{table}

\begin{table}[t]
\centering
\small
\begin{tabular}{lc}
\toprule
\textbf{Decision context} & \textbf{Accuracy (\%)} \\
\midrule
Query only & 67 \\
Single-round extracted preference & 81 \\
Multi-round extracted preference & 83 \\
\bottomrule
\end{tabular}
\caption{\apebKeepA{Effect of preference extraction on purchase identification among hard candidates.}}
\label{ap-tab:preference-extraction-validity}
\end{table}

\paragraph{\apebKeepH{Commercial-evidence audit.}}
\apebKeepH{With queries, histories, and purchase--negative pairs fixed, Table~\ref{ap-tab:commercial-evidence-ablation} contrasts full metadata, metadata without commercial attributes, and commercial-only evidence. The pattern shows that commercial cues are auxiliary, while broader product attributes provide the larger discriminative signal.}

\begin{table}[t]
\centering
\small
\setlength{\tabcolsep}{3pt}
\begin{tabularx}{\linewidth}{@{}>{\raggedright\arraybackslash}Xcc@{}}
\toprule
\textbf{\apebKeepH{Evidence}} & \textbf{\apebKeepH{GPT-5.2}} & \textbf{\apebKeepH{Gemini-2.5-Pro}} \\
\midrule
\apebKeepH{Full metadata} & \apebKeepH{83.1} & \apebKeepH{79.5} \\
\apebKeepH{Without commercial} & \apebKeepH{81.8} & \apebKeepH{78.2} \\
\apebKeepH{Commercial only} & \apebKeepH{58.2} & \apebKeepH{56.5} \\
\bottomrule
\end{tabularx}
\caption{\apebKeepH{Pairwise purchase accuracy (\%) under candidate-side evidence ablations ($N=500$); commercial denotes price, promotion, and delivery.}}
\label{ap-tab:commercial-evidence-ablation}
\end{table}

\paragraph{\apebKeepH{Human target-recoverability audit.}}
\apebKeepH{On the same 100 stratified sessions, two humans and GPT-5.2 assess randomized purchase--hard-negative pairs with identifiers, purchase status, and model outputs masked; GPT-4.1 supplies candidate-blind history summaries. Table~\ref{ap-tab:human-purchase-recoverability} reports decisive-pair recovery (92\% human agreement), while Table~\ref{ap-tab:human-hit-calibration} provides Hit@1/5 calibration under intent and refined queries, with refinement benefiting both evaluators.}

\begin{table}[t]
\centering
\small
\setlength{\tabcolsep}{4pt}
\begin{tabular}{lcc}
\toprule
\textbf{\apebKeepH{Metric}} & \textbf{\apebKeepH{Human}} & \textbf{\apebKeepH{GPT-5.2}} \\
\midrule
\apebKeepH{Pairwise purchase accuracy} & \apebKeepH{87.6\%} & \apebKeepH{83.1\%} \\
\apebKeepH{Sessions with win rate above 60\%} & \apebKeepH{95\%} & \apebKeepH{93\%} \\
\apebKeepH{Sessions with win rate above 80\%} & \apebKeepH{71\%} & \apebKeepH{63\%} \\
\bottomrule
\end{tabular}
\caption{\apebKeepH{Decisive-pair purchase recoverability on the 100-session stratified human-audit sample.}}
\label{ap-tab:human-purchase-recoverability}
\end{table}

\begin{table}[H]
\centering
\small
\begin{tabular}{lcc}
\toprule
\textbf{\apebKeepH{Query}} & \textbf{\apebKeepH{GPT-5.2 Hit@1/5}} & \textbf{\apebKeepH{Human Hit@1/5}} \\
\midrule
\apebKeepH{Intent} & \apebKeepH{25 / 79} & \apebKeepH{34 / 83} \\
\apebKeepH{Refined} & \apebKeepH{38 / 84} & \apebKeepH{40 / 86} \\
\bottomrule
\end{tabular}
\caption{\apebKeepH{GPT-5.2 and human Hit@1/5 on the same 100 stratified sessions under intent and refined queries. Values are percentages.}}
\label{ap-tab:human-hit-calibration}
\end{table}

\subsubsection{\apebKeepH{Activity and Temporal Characteristics}}
\label{ap-benchmark-validity}
\paragraph{\apebKeepH{History-activity profile.}}
\apebKeepH{\benchf targets active, multi-step shopping sessions with observable query refinement. Table~\ref{ap-tab:activity-stratification} compares the lowest and highest deciles of retained history activity; their similar performance indicates that benchmark outcomes are stable across history density.}

\begin{table}[H]
\centering
\small
\setlength{\tabcolsep}{4pt}
\begin{tabularx}{\linewidth}{@{}>{\raggedright\arraybackslash}Xc@{}}
\toprule
\textbf{\apebKeepH{History activity}} & \textbf{\apebKeepH{Avg. Hit@1 (\%)}} \\
\midrule
\apebKeepH{Bottom decile} & \apebKeepH{23.4} \\
\apebKeepH{Top decile} & \apebKeepH{23.6} \\
\bottomrule
\end{tabularx}
\caption{\apebKeepH{Average Hit@1 across evaluated models for the lowest and highest deciles of retained historical interaction count on \smartq hard-candidate cases.}}
\label{ap-tab:activity-stratification}
\end{table}

\paragraph{\apebKeepH{Search--purchase horizon.}}
\apebKeepH{\benchf uses a 30-minute search--purchase window to associate each retained session with its purchase outcome. Table~\ref{ap-tab:search-purchase-lag} compares this benchmark horizon with longer, disjoint lag bins under the same alignment and candidate-construction protocol. Alignment becomes sparser beyond 30 minutes while refined queries remain stronger throughout, characterizing the retained data as exploration-heavy, short-horizon sessions with attributable purchase outcomes.}

\begin{table}[H]
\centering
\small
\setlength{\tabcolsep}{3pt}
\begin{tabularx}{\linewidth}{@{}>{\raggedright\arraybackslash}Xcc@{}}
\toprule
\textbf{\apebKeepH{Lag}} & \textbf{\apebKeepH{Align. (\%)}} & \textbf{\apebKeepH{Hit@1 I/R (\%)}} \\
\midrule
\apebKeepH{0--30 min} & \apebKeepH{21.4} & \apebKeepH{25.6 / 37.4} \\
\apebKeepH{30--120 min} & \apebKeepH{9.2} & \apebKeepH{24.7 / 35.9} \\
\apebKeepH{2--24 h} & \apebKeepH{4.1} & \apebKeepH{23.4 / 33.8} \\
\bottomrule
\end{tabularx}
\caption{\apebKeepH{Alignment and GPT-5.2 Hit@1 across search--purchase lags using 1,000 raw pairs per bin; I/R denotes intent/refined queries.}}
\label{ap-tab:search-purchase-lag}
\end{table}

\subsubsection{\apebKeepA{Scope and Transferability}}
\label{ap-benchmark-transferability}
\apebKeepH{APeB instantiates a recurring personalization structure: recover task-relevant preferences from noisy history to resolve an underspecified request and select among plausible candidates or actions. Table~\ref{ap-tab:scope-transferability} gives representative examples of the same structure beyond product search.}

\begin{table}[H]
\centering
\small
\setlength{\tabcolsep}{4pt}
\begin{tabularx}{\linewidth}{@{}p{0.25\linewidth}X@{}}
\toprule
\textbf{\apebKeepH{Setting}} & \textbf{\apebKeepH{Example history-conditioned decision}} \\
\midrule
\apebKeepH{Mobility} & \apebKeepH{Past routes and pickup habits help resolve a brief trip request and select a route or pickup point.} \\
\midrule
\apebKeepH{Outfit recommendation} & \apebKeepH{Prior views, purchases, and occasions help resolve a vague style request and select garments or outfits.} \\
\midrule
\apebKeepH{News/content} & \apebKeepH{Reading, viewing, and skip history helps resolve a broad information need and rank articles or clips.} \\
\bottomrule
\end{tabularx}
\caption{\apebKeepH{Representative examples of APeB's core task structure beyond product search: interaction history resolves an underspecified request and guides selection among plausible options.}}
\label{ap-tab:scope-transferability}
\end{table}

\section{Experiment Setting}\label{ap-experiment-setting}

Due to architectural differences across \emph{LLM-only}, \emph{ReAct agent}, and \emph{DeerFlow agent} settings, we adopt model-specific configurations. Unless otherwise stated, all models use their default temperature. The LLM/ReAct Hit@1/5 values in Table \ref{tab:experiment-overall-query-settings} are averaged over three runs. The detailed settings are as follows:
\begin{enumerate}
    \item \textbf{Gemini-2.5-Pro} (\texttt{GEMINI} in \apebKeepA{Table~\ref{tab:experiment-overall-query-settings}}): 
    When used as a standalone LLM or within DeerFlow, we adopt the default reasoning budget. When deployed as a ReAct agent, we reduce the reasoning budget to 1024 tokens to ensure practical efficiency.
    
    \item \textbf{GPT-4.1-Mini} (\texttt{4.1MINI}): 
    All experiments use the default configuration.
    
    \item \textbf{GPT-4o}: 
    All experiments use the default configuration.
    
    \item \textbf{Qwen3-235B} (\texttt{Qwen3-235B}): 
    Although the model supports internal reasoning, we disable its thinking mode in all experiments due to interface limitations. Other settings remain default.
    
    \item \textbf{DeepSeek-R1} (\texttt{R1}): 
    Thinking mode is enabled with the default reasoning budget.
    
    \item \textbf{GPT-4.1}: 
    All experiments use the default configuration.
    
    \item \textbf{GPT-5.2}: 
    Unless specified otherwise, \texttt{GPT5.2} refers to the model configured with a minimal reasoning budget, with all other settings left unchanged.
    
    \item \textbf{GPT-5.2Think}: 
    Unless specified otherwise, 
    \apebKeepA{\texttt{GPT5.2Think} refers to the GPT-5.2 condition with the default reasoning budget, while \texttt{GPT5.2} uses the minimal reasoning-budget condition; all remaining settings follow the defaults.}

    \item \textbf{Additional model-family runs}:
    \apebKeepB{Qwen3.5, Gemini3Pro, GPT5.5, and GPT5.5Think follow the same prompt, history-selection, retrieval-tool, and parsing protocols as the corresponding main-table LLM or reasoning-model conditions.}
\end{enumerate}

\subsection{\apebKeepB{Additional Model-Family Results}}
\label{ap-experiment-additional-model-results}
\apebKeepA{Table~\ref{ap-tab:additional-model-family-results} reports additional model-family results using the same query, method, and metric structure as Table~\ref{tab:experiment-overall-query-settings}.}

\begin{table*}[t]
\centering
\small
\setlength{\tabcolsep}{3pt}
\renewcommand{\arraystretch}{0.96}
\resizebox{\textwidth}{!}{%
\begin{tabular}{lllcccc}
\toprule
\textbf{Query} & \textbf{Method} & \textbf{Metric (\%)} & \textbf{Qwen3.5} & \textbf{Gemini3Pro} & \textbf{GPT5.5} & \textbf{GPT5.5Think} \\
\midrule
\multirow{4}{*}{\textbf{\smartqB}}
& \multirow{2}{*}{\textbf{Single-Prompt}} & Hit@1 & 22.9 & 24.1 & 25.4 & 25.0 \\
& & Hit@5 & 72.4 & 76.7 & 78.1 & 78.2 \\
\cmidrule(lr){2-7}
& \multirow{2}{*}{\textbf{ReAct}} & Hit@1 & 24.4 & 24.8 & 26.6 & 26.5 \\
& & Hit@5 & 71.3 & 76.9 & 78.6 & 78.1 \\
\midrule
\multirow{4}{*}{\textbf{\matchqB}}
& \multirow{2}{*}{\textbf{Single-Prompt}} & Hit@1 & 36.1 & 35.1 & 37.2 & 37.3 \\
& & Hit@5 & 81.9 & 83.2 & 84.2 & 84.4 \\
\cmidrule(lr){2-7}
& \multirow{2}{*}{\textbf{ReAct}} & Hit@1 & 37.8 & 37.1 & 43.2 & 43.0 \\
& & Hit@5 & 82.1 & 83.5 & 84.8 & 84.5 \\
\bottomrule
\end{tabular}
}
\caption{\apebKeepH{Additional model-family Hit@1/5 results under the same intent/refined hard-candidate protocol as Table~\ref{tab:experiment-overall-query-settings}. Values are percentages.}}
\label{ap-tab:additional-model-family-results}
\end{table*}

\subsubsection{\texorpdfstring{\apebKeepH{Controlled Qwen Scale Study}}{Controlled Qwen Scale Study}}
\label{ap-experiment-qwen-scale}
\apebKeepH{To isolate model scale, we evaluate five Qwen models on the same stratified sample of 500 cases drawn from the full 5,648-case benchmark, fixing hard candidates and 60-record category-prior histories across models. All results use Single-Prompt inference and three runs. Table~\ref{ap-tab:qwen-scale-study} provides the controlled comparison behind the main-text finding: scaling improves absolute Hit@1 without eliminating the intent--refined gap.}

\begin{table}[H]
\centering
\small
\setlength{\tabcolsep}{4pt}
\begin{tabularx}{\linewidth}{@{}>{\raggedright\arraybackslash}p{0.29\linewidth}*{4}{>{\centering\arraybackslash}X}@{}}
\toprule
\textbf{\apebKeepH{Model}} & \multicolumn{2}{c}{\textbf{\apebKeepH{Hit@1}}} & \multicolumn{2}{c}{\textbf{\apebKeepH{Rubric mean}}} \\
\cmidrule(lr){2-3}\cmidrule(l){4-5}
& \textbf{\apebKeepH{I}} & \textbf{\apebKeepH{R}} & \textbf{\apebKeepH{I}} & \textbf{\apebKeepH{R}} \\
\midrule
\apebKeepH{Qwen3.5-4B}  & \apebKeepH{18.2} & \apebKeepH{26.1} & \apebKeepH{52.0} & \apebKeepH{67.3} \\
\apebKeepH{Qwen3.5-9B}  & \apebKeepH{18.9} & \apebKeepH{28.8} & \apebKeepH{53.7} & \apebKeepH{73.0} \\
\apebKeepH{Qwen3.5-27B} & \apebKeepH{20.6} & \apebKeepH{31.9} & \apebKeepH{57.3} & \apebKeepH{79.7} \\
\midrule
\apebKeepH{Qwen3-32B}   & \apebKeepH{20.4} & \apebKeepH{33.9} & \apebKeepH{56.7} & \apebKeepH{76.3} \\
\apebKeepH{Qwen3-235B}  & \apebKeepH{22.8} & \apebKeepH{35.9} & \apebKeepH{58.3} & \apebKeepH{78.3} \\
\bottomrule
\end{tabularx}
\caption{\apebKeepH{Controlled Qwen scaling on the shared 500-case sample from \benchs. Rubric mean averages QI, PQ, and RQ; all values are three-run percentages, with I/R denoting intent/refined queries.}}
\label{ap-tab:qwen-scale-study}
\end{table}

\subsection{\texorpdfstring{\apebKeepH{Statistical Reliability}}{Statistical Reliability}}
\label{ap-statistical-reliability}
\apebKeepH{Table~\ref{ap-tab:paired-bootstrap} quantifies the marginal \smartq-query gains reported in the main text. After averaging three runs per session--method--backbone cell, we resample session identifiers 10,000 times with method and backbone pairing preserved, then macro-average differences from Single-Prompt equally across eight backbones. ReAct's 95\% interval excludes zero only at Hit@1, whereas \ouragent's intervals exclude zero at both Hit@1 and Hit@5.}

\begin{table}[t]
\centering
\small
\setlength{\tabcolsep}{4pt}
\begin{tabularx}{\linewidth}{@{}>{\raggedright\arraybackslash}p{0.20\linewidth}*{2}{>{\centering\arraybackslash}X}@{}}
\toprule
\textbf{\apebKeepH{Method}} & \multicolumn{2}{c}{\textbf{\apebKeepH{Difference from Single-Prompt (pp)}}} \\
\cmidrule(l){2-3}
& \textbf{\apebKeepH{Hit@1 [95\% CI]}} & \textbf{\apebKeepH{Hit@5 [95\% CI]}} \\
\midrule
\apebKeepH{ReAct} & \apebKeepH{+1.64 [+0.7, +2.6]} & \apebKeepH{+0.36 [-0.4, +1.1]} \\
\midrule
\apebKeepH{VQRA} & \apebKeepH{+1.70 [+0.8, +2.6]} & \apebKeepH{+1.08 [+0.3, +1.8]} \\
\bottomrule
\end{tabularx}
\caption{\apebKeepH{Paired session-level bootstrap macro differences for Intent hard candidates, using three-run averages, equal weighting over eight backbones, and 10,000 resamples.}}
\label{ap-tab:paired-bootstrap}
\end{table}

\subsection{\apebKeepA{Additional Baseline Settings}}
\label{ap-experiment-additional-baselines}
\apebKeepH{We organize the additional comparisons into embedding/reranking baselines and memory-agent workflows.}

\subsubsection{\apebKeepH{Embedding and Reranking Baselines}}
\label{ap-experiment-embedding-reranking-baselines}
\apebKeepA{We choose UniSAR~\cite{pps-s-6} as the primary non-LLM PPS/SAR baseline because it is publicly available and directly models user transition behavior between search and recommendation.}
\apebKeepH{Table~\ref{ap-tab:additional-baselines} reports reproducible OpenAI and Qwen3 embedding/reranking checks~\cite{qwenembedding} under the shared \smartq hard-candidate protocol.}

\begin{table}[t]
\centering
\small
\setlength{\tabcolsep}{4pt}
\begin{tabularx}{\linewidth}{@{}>{\raggedright\arraybackslash}Xcc@{}}
\toprule
\textbf{\apebKeepH{Baseline}} & \multicolumn{2}{c}{\textbf{\apebKeepH{Hit@1 (\%)}}} \\
\cmidrule(l){2-3}
& \textbf{\apebKeepH{No history}} & \textbf{\apebKeepH{With history}} \\
\midrule
\apebKeepH{OpenAI embedding} & \apebKeepH{15.1} & \apebKeepH{13.9} \\
\midrule
\apebKeepH{Qwen3-Embedding-8B} & \apebKeepH{17.7} & \apebKeepH{18.0} \\
\midrule
\apebKeepH{Qwen3-Reranker-4B} & \apebKeepH{9.0} & \apebKeepH{14.3} \\
\bottomrule
\end{tabularx}
\caption{\apebKeepH{Additional embedding and reranking checks on \smartq hard candidates under the shared candidate protocol.}}
\label{ap-tab:additional-baselines}
\end{table}

\apebKeepA{For embedding baselines, query and candidate text are encoded and ranked by similarity; the history-conditioned variant appends selected user-history evidence before scoring.}
\apebKeepA{For trained embedding and reranker baselines, training uses the retained search-purchase supervision available to the baseline protocol, and evaluation uses the \benchs hard-candidate set.}
\apebKeepA{The baselines slightly differ in signal access. UniSAR and the trained embedding/reranker variants learn from historical search-purchase supervision, whereas LLM prompting and agent workflows operate primarily through in-context history \apebKeepB{evidence}.}
\apebKeepA{UniSAR also uses cross-session negatives, which is standard in SAR/PPS training \cite{onerecthink}.
}

\subsubsection{\apebKeepH{Memory-Agent Baselines}}
\label{ap-experiment-memory-agent-baselines}
\apebKeepH{Mem0 and MemAgent augment an evaluated LLM backbone with an explicit memory component \cite{Mem0, memagent}. Table~\ref{ap-tab:memory-agent-results} reports their results using an LLM mix comparable to Table~\ref{tab:experiment-overall-query-settings}.}

\begin{table*}[t]
\centering
\small
\setlength{\tabcolsep}{3pt}
\renewcommand{\arraystretch}{0.96}
\resizebox{\textwidth}{!}{%
\begin{tabular}{lllcccccc}
\toprule
\textbf{Query} & \textbf{Workflow} & \textbf{Metric (\%)}
& \textbf{Gemini3Pro} & \textbf{4.1mini} & \textbf{GPT4o} & \textbf{Qwen3.5} & \textbf{GPT4.1} & \textbf{GPT5.5} \\
\midrule
\multirow{4}{*}{\textbf{\smartqB}}
& \multirow{2}{*}{\textbf{Mem0}} & Hit@1 & 26.1 & 25.5 & 25.9 & 25.1 & 25.8 & 24.1 \\
& & Hit@5 & 78.5 & 73.8 & 76.5 & 74.9 & 77.1 & 76.9 \\
\cmidrule(lr){2-9}
& \multirow{2}{*}{\textbf{MemAgent}} & Hit@1 & 23.8 & 23.7 & 24.2 & 22.5 & 25.0 & 24.0 \\
& & Hit@5 & 76.5 & 71.5 & 75.4 & 70.5 & 76.1 & 77.1 \\
\midrule
\multirow{4}{*}{\textbf{\matchqB}}
& \multirow{2}{*}{\textbf{Mem0}} & Hit@1 & 34.3 & 35.4 & 37.5 & 37.9 & 38.2 & 38.1 \\
& & Hit@5 & 81.5 & 78.9 & 83.9 & 82.1 & 82.9 & 82.1 \\
\cmidrule(lr){2-9}
& \multirow{2}{*}{\textbf{MemAgent}} & Hit@1 & 34.2 & 35.8 & 37.3 & 34.2 & 38.6 & 37.9 \\
& & Hit@5 & 82.0 & 79.1 & 80.2 & 79.9 & 82.8 & 82.5 \\
\bottomrule
\end{tabular}
}
\caption{\apebKeepH{Mem0 and MemAgent Hit@1/5 results across six backbones under intent and refined hard-candidate evaluation. Values are percentages.}}
\label{ap-tab:memory-agent-results}
\end{table*}

\subsection{Single-Prompt LLM}\label{ap-experiment-single-prompt}
\apebKeepB{This subsection describes the shared direct-prompt protocol; agent-specific extensions are described in Appendix~\ref{ap-experiment-agent} and Appendix~\ref{ap-deerflow}.}

\subsubsection{LLM Prompt}\label{ap-experiment-single-prompt-prompt}

For all LLMs evaluated under the same setting, we use an identical system prompt to ensure fair comparison. The prompt is instantiated according to the scenario configuration. For example, when the input consists of a \smartq\ query with hard negative candidates, the system prompt takes the following form (details omitted for brevity):

\begin{prompt}
TASK: Structured Product Recommendation (Exactly 5 Items)

Your goal is to recommend exactly five products that best match the user's explicit needs, using only data from:

  (1) $<$history$>$ --- chronological user interactions with indices (action type, product in this history, price)
  
  (2) $<$query$>$ --- the user's current query (may be ambiguous or intentional)
  
  (3) $<$candidates$>$ --- product candidates (index, title, price)

\promptsep

STEP 0 --- Determine the User's Explicit Target (Ambiguous Query Handling)

\promptsep

(...)

\promptsep

STEP 1 --- Identify Important History Entries Based on the Query

\promptsep

(...)

\promptsep

STEP 2 --- Extract Explicit Features (Summarized) \& Rank Relevance

\promptsep

(...)

\promptsep

STEP 3 --- Evaluate All Candidate Products

\promptsep

(...)

\promptsep

STEP 4 --- Produce Exactly 5 Final Recommendations

\promptsep

(...)

\promptsep

OUTPUT SPECIFICATION (STRICT JSON ONLY)

\promptsep

Output raw JSON conforming to the \texttt{Recommendation} interface \textbf{without} wrapping it in triple back-ticks.  

Type schema:

interface Product \{

    rank: int;                                // 1--5, unique ascending
    
    index: int;                               // candidate index
    
    product\_title: string;                    // rewritten title (no new facts)
    
    product\_description: string;              // rewritten description
    
    product\_recommendation\_reason: List[string];  // grounded in query + Step 0 + summarized\_features
    
\}

interface SupportingHistoryFeature \{

    summarized\_feature: string;
    
    support\_history\_index: List[int];
    
    relevance\_rank: int;                 // from Step 2
    
    support\_product: List[int];          // candidate indices supported
    
\}

interface Recommendation \{

    target\_product: string;              // derived in Step 0
    
    concerned\_features: string;          // derived in Step 2
    
    products: List[Product];             // exactly 5 items
    
    histories: List[SupportingHistoryFeature];   // $>$5 entries, sorted by relevance\_rank
    
\}
End schema.

\end{prompt}

In other scenarios, the system prompt may undergo minor specification changes, while its overall structure remains unchanged.

The \texttt{<history>}, \texttt{<query>}, and \texttt{<candidates>} are supplied in the user prompt using a fixed template. For each interaction, to avoid introducing irrelevant context, only essential fields---interaction type, title, and price---are included; all identifiers and sensitive attributes are encrypted.
\apebKeepH{For presentation, entity names are replaced with placeholders, and prices are slightly perturbed while preserving their scale.}

\begin{prompt}
$<$history$>$

\apebKeepH{1. Action type: order product; Product name: [APPROVED BY A SKINCARE LECTURER] [BRAND-A] Water Gel Moisturizer 30g | Brightens \& Reduces Dark Spots | Maintains Moisture for 48 Hours, Strengthens the Skin Barrier | Helps Reduce Acne and Soothe Redness for Glowing Skin | Contains Niacinamide and Vitamin C; Price: 3.9 usd}

(...)

$<$/history$>$

$<$query$>$

\apebKeepH{dd cream [BRAND-B]}

$<$/query$>$

$<$candidates$>$

\apebKeepH{1. Product name: (DRY SKIN) [BRAND-C] Emulsion for DRY SKIN - Moisturizes and firms dry skin; Price: 15.2 usd}

(...)

$<$/candidates$>$

\end{prompt}

Finally, we parse model outputs using a Pydantic-based parser and compute standard retrieval metrics, including HR@1 and HR@5.

\subsubsection{History Record Selection}\label{ap-experiment-llm-selection-history}
To control prompt length while preserving essential interaction semantics (e.g., action type, product title, and price), we limit the history to at most 60 records using an importance-aware selection mechanism. 
\apebKeepA{Results in Table~\ref{tab:experiment-overall-query-settings} primarily adopt the category-prior strategy using the category label associated with the provided intent query.}

\paragraph{Importance scoring.}
Each interaction is assigned an importance score determined by interaction type and dwell time, with stronger behavioral signals ranked as
\(\texttt{order} > \texttt{cart} > \texttt{like} > \texttt{view}\).
For interactions of the same type and category, longer dwell time yields higher importance.

\paragraph{Time-prior selection.}
From the full interaction history, we first retain records whose importance exceeds a predefined ratio threshold. The final 60 records are then selected by recency within this importance-filtered subset, ensuring that retained histories emphasize both behavioral salience and temporal relevance.

\paragraph{\apebKeepA{Category-prior selection.}}
\apebKeepA{We first assign the provided intent query to a coarse shopping category label, such as skin care, toys, or clothing, following the category taxonomy summarized in Appendix~\ref{ap-benchmark-stat}. We then include interactions whose category matches this intent-query category, up to 60 records, ranked by importance. If fewer than 60 same-category records are available, the remaining slots are filled using the time-prior selection strategy above. This approach maximizes query-relevant category evidence while retaining salient and recent interactions.}
\apebKeepA{This schema does not use the purchased item or reveal target-product information.}
\apebKeepA{We include this variant to test whether query-relevant category histories help models extract user preferences from noisy behavior records.}

\subsubsection{\apebKeepA{History Context Stress Tests}}
\label{ap-experiment-history-stress}
\apebKeepA{In Table \ref{ap-tab:history-stress-tests}, we run additional stress tests on 500 randomly selected \smartq hard-candidate sessions to isolate whether history gains come from useful user evidence.}
\apebKeepA{Shuffling histories reduces Hit@1 for both R1 and Qwen3, indicating that performance depends on coherent user-specific evidence.}
\apebKeepA{Modality-specific results show that product and media/video histories both provide preference signals, while increasing the context from 60 to 240 records does not improve performance.}
\apebKeepA{These results demonstrate that useful history selection and grounding matter more than simply exposing longer histories.}

\begin{table}[t]
\centering
\small
\setlength{\tabcolsep}{3pt}
\renewcommand{\arraystretch}{0.98}
\begin{tabularx}{\linewidth}{@{}>{\raggedright\arraybackslash}p{0.20\linewidth}>{\raggedright\arraybackslash}Xcc@{}}
\toprule
\textbf{\apebKeepH{Test}} & \textbf{\apebKeepH{History input}} & \textbf{\apebKeepH{R1}} & \textbf{\apebKeepH{Qwen3}} \\
\midrule
\multirow{2}{*}{\apebKeepH{Assignment}} & \apebKeepH{True history} & \apebKeepH{24.2} & \apebKeepH{22.7} \\
\cmidrule(l){2-4}
& \apebKeepH{Shuffled history} & \apebKeepH{21.9} & \apebKeepH{20.5} \\
\midrule
\multirow{2}{*}{\apebKeepH{Modality}} & \apebKeepH{Product only} & \apebKeepH{24.6} & \apebKeepH{22.7} \\
\cmidrule(l){2-4}
& \apebKeepH{Media/video only} & \apebKeepH{24.1} & \apebKeepH{22.3} \\
\midrule
\multirow{2}{*}{\apebKeepH{Length}} & \apebKeepH{60 records} & \apebKeepH{24.2} & \apebKeepH{22.7} \\
\cmidrule(l){2-4}
& \apebKeepH{240 records} & \apebKeepH{23.8} & \apebKeepH{21.9} \\
\bottomrule
\end{tabularx}
\caption{\apebKeepH{History-context stress tests on 500 sampled Intent hard-candidate sessions. Values are Hit@1 (\%) for DeepSeek-R1 and Qwen3-235B; only the history input varies.}}
\label{ap-tab:history-stress-tests}
\end{table}

\paragraph{\apebKeepH{Explicit preference representation.}}
\apebKeepH{On 500 randomly selected Intent hard-candidate sessions, a GPT-5.2 extractor converts the query and raw pre-session history into explicit preferences with supporting history items. Candidate items and purchase status are masked from the extractor, and the resulting preferences replace the raw history in the Single-Prompt decision input.}
\apebKeepH{Explicit preferences increase Hit@1 performances for GPT-4.1-mini and Qwen3, while GPT-5.2 barely changes. This backbone-controlled comparison complements the staged extraction diagnostic in Table~\ref{ap-tab:preference-extraction-validity}.}

\begin{table}[t]
\centering
\small
\setlength{\tabcolsep}{4pt}
\begin{tabularx}{\linewidth}{@{}>{\raggedright\arraybackslash}Xcc@{}}
\toprule
\textbf{\apebKeepH{Backbone}} & \multicolumn{2}{c}{\textbf{\apebKeepH{Hit@1 (\%)}}} \\
\cmidrule(l){2-3}
& \textbf{\apebKeepH{Raw history}} & \textbf{\apebKeepH{Extracted prefs.}} \\
\midrule
\apebKeepH{GPT-4.1-mini} & \apebKeepH{23.2} & \apebKeepH{24.4} \\
\midrule
\apebKeepH{Qwen3-235B} & \apebKeepH{22.9} & \apebKeepH{24.8} \\
\midrule
\apebKeepH{GPT-5.2} & \apebKeepH{25.4} & \apebKeepH{25.3} \\
\bottomrule
\end{tabularx}
\caption{\apebKeepH{Hit@1 (\%) with raw histories versus extracted preferences across three backbones on 500 sampled Intent hard-candidate sessions.}}
\label{ap-tab:explicit-preference-representation}
\end{table}

\subsubsection{\apebKeepA{Contamination and Counterfactual History Check}}
\label{ap-experiment-contamination-counterfactual}
\apebKeepA{
We also \apebKeepB{probe} \textit{data contamination} through a memorization hypothesis: if decisions mainly rely on memorized query-item associations, then replacing the user's real history with unrelated counterfactual histories should have limited effect. We therefore hold the query and candidate set fixed and compare query-only, counterfactual-history, and real-history contexts on 500 sampled sessions.}
\apebKeepA{For each sampled session, the query and candidate set remain unchanged. In the query-only condition, the model receives no historical records. In the counterfactual condition, the true user's history is replaced with a length-matched history sampled from a different user, excluding records from the target user and session. In the real-history condition, the model receives the original user history under the same formatting and parsing protocol. We report pairwise preference accuracy in Table \ref{ap-tab:counterfactual-history-contamination}: the percentage of purchase-vs-hard-negative comparisons in which the model selects the observed purchased item.}

\begin{table}[t]
\centering
\small
\setlength{\tabcolsep}{3pt}
\begin{tabularx}{\linewidth}{@{}>{\raggedright\arraybackslash}Xccc@{}}
\toprule
\textbf{\apebKeepH{Model}} & \multicolumn{3}{c}{\textbf{\apebKeepH{Pairwise accuracy (\%)}}} \\
\cmidrule(l){2-4}
& \textbf{\apebKeepH{Query only}} & \textbf{\apebKeepH{Ctf.}} & \textbf{\apebKeepH{Real}} \\
\midrule
\apebKeepH{GPT-4.1} & \apebKeepH{68.3} & \apebKeepH{67.7} & \apebKeepH{79.3} \\
\midrule
\apebKeepH{GPT-5.2} & \apebKeepH{70.8} & \apebKeepH{68.1} & \apebKeepH{83.1} \\
\midrule
\apebKeepH{Gemini-2.5-Pro} & \apebKeepH{69.1} & \apebKeepH{68.5} & \apebKeepH{79.5} \\
\bottomrule
\end{tabularx}
\caption{\apebKeepH{Pairwise preference accuracy for selecting the observed purchase over hard negatives on 500 sampled sessions. Query and candidates are fixed; Ctf. uses a length-matched history from another user.}}
\label{ap-tab:counterfactual-history-contamination}
\end{table}

\apebKeepA{Counterfactual histories stay close to query-only performance for all three models, while real histories improve over counterfactual histories substantially. This pattern suggests that useful performance depends on the matching user's context rather than arbitrary additional history text. The check provides evidence that memorized query-item associations are unlikely to be the dominant source of the observed history-conditioned gains.}
\apebKeepA{\apebKeepB{To address possible leakage concerns during data collection and evaluation, Appendix}~\ref{ap-experiment-llm-role-audit} separately audits GPT-4.1's construction and diagnostic roles; GPT-4.1 does not determine Hit@K labels, and its \apebKeepB{evaluated task} performance is not superior to \apebKeepB{the other} evaluated models.}

\subsection{ReAct Agent}\label{ap-experiment-agent}

Our ReAct agent is implemented using the open-source LangChain framework. Compared to the single-prompt LLM setting, the system prompt follows the same high-level structure but additionally specifies the available tools and their usage constraints.

\begin{prompt}
\promptsep

AVAILABLE TOOL 1 --- structured\_search\_tool

\promptsep

Purpose:

Use this tool only to retrieve literal factual clarifications about products or history entries.

Permitted data sources:

- Candidate product detailed descriptions  

- History product detailed descriptions  

- Candidate product detailed features  

- History product detailed features  

- History video descriptions (OCR/ASR)

Tool Query Rules:

1. You must generate your own search queries from:

   - literal terms in the user query  
   
   - literal terms relevant to target\_product or concerned\_features  
   
   - literal candidate titles or features  
   
   - literal history text  
   
   - interpretation-critical terms ONLY when clarifying explicit text

4. You may only rely on:

   - explicit user inputs  
   
   - explicit tool results  

\promptsep

AVAILABLE TOOL 2 --- web\_search\_tool

\promptsep

(...)
\end{prompt}

Beyond the explicit system-prompt specification, model providers (e.g., GPT and Gemini) impose additional constraints and instructions on tool invocation through their respective interfaces. We strictly follow the official usage guidelines of each provider. The user prompt is identical to that used in the single-prompt LLM setting.

\paragraph{Available Tools.}
\begin{enumerate}
    \item \textbf{Structured Search Tool.} 
    In addition to item titles, \benchf provides rich auxiliary information, including product descriptions, video OCR/ASR transcripts, and structured attributes. As these contents exceed prompt-length limits, they are organized in a fixed schema and stored in a retrieval database, which the agent queries via this tool (see Appendix~\ref{ap-retrieval-database}).
    
    \item \textbf{Web Search Tool.} 
    We wrap the web search API provided by YOU\footnote{https://you.com/home} as a callable tool. Each query returns the top-3 relevant text chunks.
    
    \item \textbf{Crawl Tool.} 
    We use the Jina AI crawling API\footnote{https://jina.ai/} to fetch web pages and convert them into agent-friendly Markdown representations.
    
    \item \textbf{Code Execution Tool.} 
    We employ the Python REPL tool from LangChain\footnote{https://www.langchain.com/}, enabling the agent to execute code and observe intermediate results.
\end{enumerate}

\subsubsection{\apebKeepA{ReAct Failure-Source Analysis}}
\label{ap-react-failure-analysis}
\apebKeepA{We analyze logged ReAct traces under the same hard-candidate evaluation setting used in the main tables. In Table \ref{ap-tab:react-tool-utility}, for each trace, we report the number of reasoning rounds reached, and the AC/QC/QH/QQ diagnostic scores defined in Appendix~\ref{ap-experiment-rubric-aspect}. Retrieval quality is measured by rubric-scored relevance to the issued query and utility to the user intent of the retrieved evidence.}

\begin{table}[t]
\centering
\small
\begin{tabular}{lcc}
\toprule
\textbf{Metric} & \textbf{\smartqB} & \textbf{\matchqB} \\
\midrule
Candidate retrieval relevance (QC-1) & 88 & 92 \\
History retrieval relevance (QH-1) & 81 & 86 \\
Candidate retrieval utility (QC-2) & 68 & 80 \\
History retrieval utility (QH-2) & 49 & 55 \\
Query total quantity (QQ) & 8.1 & 8.5\\ 
\bottomrule
\end{tabular}
\caption{
\apebKeepA{ReAct retrieval diagnostics: relevance and utility of retrieved candidate/history evidence, plus total query count.}
}
\label{ap-tab:react-tool-utility}
\end{table}

\apebKeepA{The reasonable query quantity indicates that the ReAct degradation is not primarily due to agents failing to call the available tools. The lower QH scores indicate that the harder bottleneck is formulating history-relevant queries under vague goals: agents often produce locally plausible search terms, but these terms retrieve weaker personalization evidence than candidate metadata.}

\begin{table}[t]
\centering
\small
\begin{tabular}{lcc}
\toprule
\textbf{Round} & \textbf{Trace ratio (\%)} & \textbf{AC} \\
\midrule
First round & 100.0 & 63 \\
Second round & 13.6 & 61 \\
Third round & 1.7 & 65 \\
\bottomrule
\end{tabular}
\caption{\apebKeepA{Round-depth and coherence for ReAct traces. Ratio is the percentage of traces reaching each round; AC is the mean agent-coherence score for that round.}}
\label{ap-tab:react-round-ac}
\end{table}

\apebKeepA{Table \ref{ap-tab:react-round-ac} shows that most traces stop after one round, and the small subset that continues does not show improved coherence. This pattern argues against over-exploration as the main cause of degradation and indicates limited spontaneous self-correction.}

\begin{table}[t]
\centering
\small
\begin{tabular}{lcc}
\toprule
\textbf{Model} & \textbf{Vanilla ReAct} & \textbf{Self-correction prompt} \\
\midrule
Qwen3 & 24.6 & 24.7 \\
GPT-5.2 & 26.2 & 26.5 \\
R1 & 24.8 & 25.3 \\
Gemini & 24.2 & 24.6 \\
GPT-4.1 & 25.8 & 25.2 \\
\bottomrule
\end{tabular}
\caption{\apebKeepA{Self-correction prompt ablation for ReAct on \smartq hard-candidate cases. Values are Hit@1 (\%).}}
\label{ap-tab:react-self-correction}
\end{table}

\apebKeepA{We additionally report ReAct \apebKeepB{results} with the explicit self-correction instruction prompt in Table \ref{ap-tab:react-self-correction}. This produces \apebKeepB{only} small\apebKeepB{, model-dependent} changes across models.}
\apebKeepA{In Table \ref{ap-tab:react-backend-sensitivity}, we report Hit@1 results when varying the retrieval backend. The minimal differences suggest that backend choice alone does not explain the ReAct gap. These checks indicate that a redesigned agent architecture or task-specific planner may be needed to improve personalization performance.}

\begin{table}[t]
\centering
\small
\setlength{\tabcolsep}{3pt}
\begin{tabularx}{\linewidth}{@{}>{\raggedright\arraybackslash}Xccc@{}}
\toprule
\textbf{\apebKeepH{Backend}} & \multicolumn{3}{c}{\textbf{\apebKeepH{ReAct Hit@1 (\%)}}} \\
\cmidrule(l){2-4}
& \textbf{\apebKeepH{Qwen3}} & \textbf{\apebKeepH{R1}} & \textbf{\apebKeepH{GPT-4.1}} \\
\midrule
\apebKeepH{BM25} & \apebKeepH{24.6} & \apebKeepH{24.8} & \apebKeepH{25.8} \\
\midrule
\apebKeepH{OpenAI embedding} & \apebKeepH{24.7} & \apebKeepH{25.3} & \apebKeepH{25.2} \\
\bottomrule
\end{tabularx}
\caption{\apebKeepH{ReAct retrieval-backend sensitivity on Intent hard-candidate cases. Backbones are Qwen3-235B, DeepSeek-R1, and GPT-4.1.}}
\label{ap-tab:react-backend-sensitivity}
\end{table}

\subsection{DeerFlow Agent System}\label{ap-deerflow}

\begin{figure*}[t]
    \centering
    \includegraphics[width=0.8\textwidth]{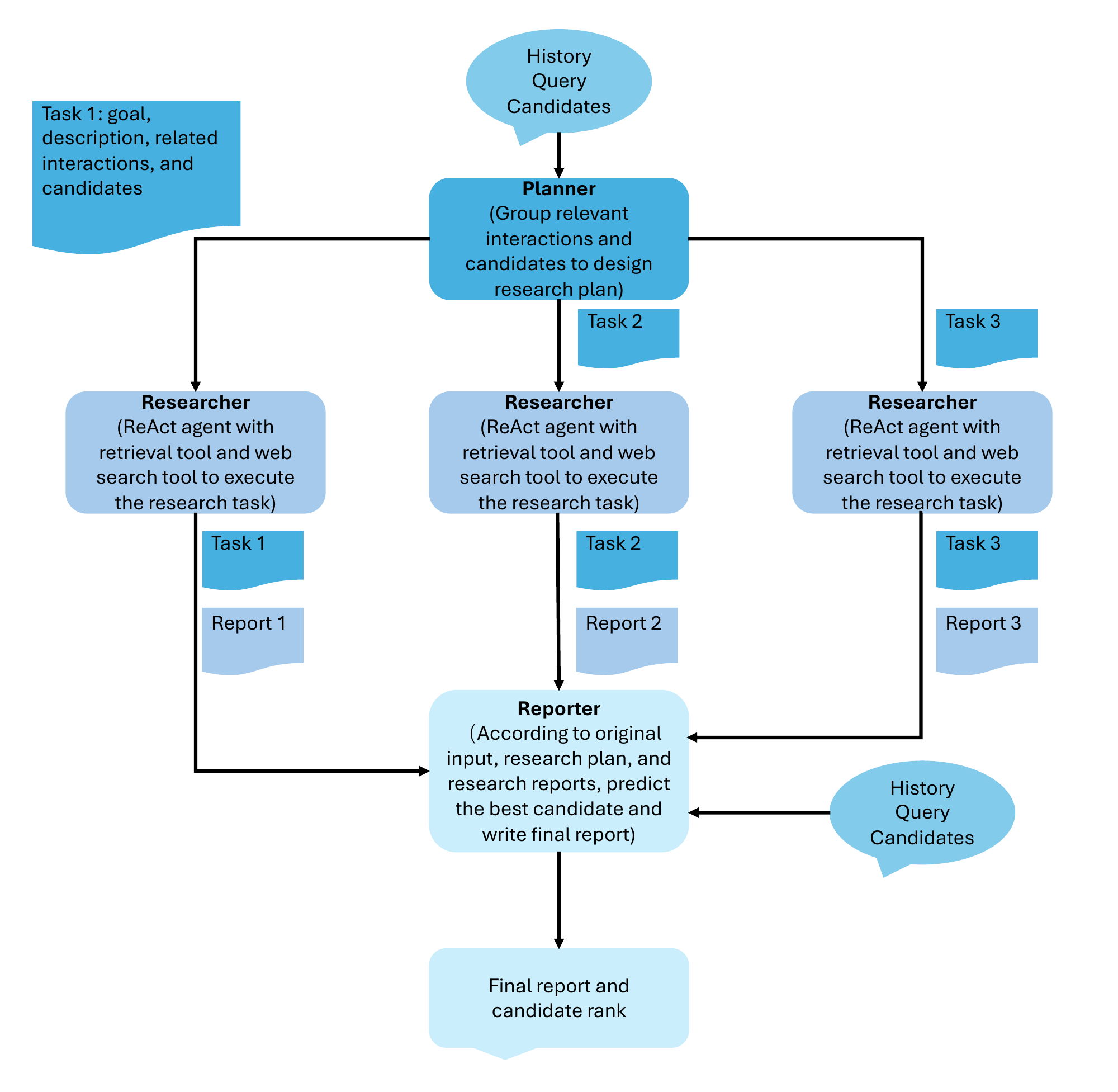}%
    \caption{Overview of our modified DeerFlow agent system.}
    \label{fig:ap-deerflow}
\end{figure*}

As the original DeerFlow agent system includes components unsuitable for \benchf\ evaluation (e.g., coordinators and human feedback loops), we simplify its architecture to the workflow illustrated in Figure~\ref{fig:ap-deerflow}. For all DeerFlow-based experiments, the evaluated model simultaneously assumes the roles of \emph{Planner}, \emph{Researcher}, and \emph{Reporter}.

\subsubsection{Planner}

The Planner receives the user input---consisting of history, query, and candidates---directly as input. It is instructed to group relevant historical interactions and candidate items, and to formulate a set of research tasks aimed at information gathering and user preference investigation. The system prompt follows the form:

\begin{prompt}
(...)

Follow these steps to generate a research plan:

1. \textbf{Understand query and products} - Understand the action content. Reason about the hidden pattern. Catch the relation between the action content, search query, and candidates. Based on your thinking result, infer a grouping strategy that is helpful for researcher to dig out the important and detailed features.

2. \textbf{Group products} - Based on the grouping strategy derived from step 1, break the history interactions and candidates into different groups. In each groups, please clearly list the index of the interactions and products in this group, but you don't need to include the interaction content or product title in the task description because it will be automatically passed to researcher based on your index output.

3. \textbf{Check group size} - Consider whether the size of each group is reasonable to form a research task with medium difficulty and clear scope (a desired size will be between 6 and 10 interactions and candidates related with each other). If not, adjust the grouping strategy.

4. \textbf{Formulate research goal} - Formulate a clear and concise research goal for each group. The goal should be suitable enough to utilize the ability of researcher (they are strong models with retrieval tools).

(...)

Output raw JSON conforming to the \texttt{Plan} interface \textbf{without} wrapping it in triple back-ticks.  

Type schema:

interface Step \{

  need\_search: bool; // Must be explicitly set for each step
  
  title: str;
  
  description: str; // Specify exactly the product link and title \apebKeepH{belonging to this group}, and discuss what features to collect from these products.
  
  interactions\_index\_list: List[int]; // The index of the products in this group
  
  candidates\_index\_list: List[int]; // The index of the candidates in this group
  
\}

interface Plan \{
  
  has\_enough\_context: bool;
  
  thought: str;
  
  title: str;
  
  steps: List[Step]; // Research \& Processing steps to get more context
  
\}

End schema.

(...)
\end{prompt}

\subsubsection{Researcher}

The Researcher agent shares the same implementation and toolset as the ReAct agent described in Section~\ref{ap-experiment-agent}. The only differences lie in the input and system prompt: the Researcher receives the research tasks generated by the Planner, and operates under the following system prompt:
\begin{prompt}
\# Task Background

You are a \textbf{research agent} in an e-commerce recommendation system. This recommendation system is responsible for predicting what user will order for current search query based on the history interactions list and the candidate list of products. It made up of three agents:

1. \textbf{Planning agent} - Understand the user's search query, group the history interactions and candidates by feature or category, and break the overall task into sub-tasks on different group of products for the research agent.

2. \textbf{Research agent (you)} -  Uses various retrieval tools to explore the product information in depth for the allocated group and analyze the feature of this group that user might be interested in or not interested in.

3. \textbf{Reporter agent} - Compile the results from the research agent and combine it with user's search query to generate a final report on how much is a candidate suitable for user and what is the most suitable candidate for user.

You are dedicated to conducting in-depth investigations about part of user history interactions and product candidates. You will be provided with a research topic, and a research description, together with \apebKeepH{a carefully selected and grouped set of history interactions and candidates on which this research topic focuses}.

(...)

\# Output Format

- Provide a structured response in markdown format.

- Include the following sections:

    - \textbf{Problem Statement}: Restate the problem for clarity.
    
    - \textbf{Findings}: Organize your findings by topic rather than by tool used. For each major finding:
    
        - Summarize the key information
        
        - Track the sources of information but DO NOT include inline citations in the text
        
    - \textbf{Conclusion}: Provide a synthesized response to the problem based on the gathered information.

(...)
\end{prompt}

\subsubsection{Reporter}

The Reporter agent receives the original user input, the research plan, and all research reports produced by the Researchers. It is instructed to synthesize the collected evidence into a final report and produce a ranked list of candidates. The system prompt takes the following form:

\begin{prompt}
(...)

You should analyze the research result, user persona, and user history following these steps:

1. List: In the research result of different groups, there are lots of features that are unique or similar to each other. Please distinguish them and list them out clearly. For those similar features, you can combine them if they share exactly the same meaning, but **NEVER** combine them into one general feature because it will make the report too general and not able to meet user's needs.

2. Ranking: Based on user persona and research results, rank the primary features to consider first when \apebKeepB{the} user \apebKeepB{issues} such query content. Notice that common features are not \apebKeepB{necessarily} to be ranked higher, because details always \apebKeepB{matter} to the user.

3. Choosing: Based on the user persona and ranking result, discuss why each candidate is or is not suitable for \apebKeepB{the} user, and present the most suitable candidate for \apebKeepB{the} user. Finally, you should output a rank for all \apebKeepB{candidates} based on how much it is suitable for \apebKeepB{the} user.

(...)

Output raw JSON conforming to the \texttt{Recommendation} interface \textbf{without} wrapping it in triple back-ticks.  

Type schema:

interface Product \{
    product\_title: str; // The title of the product
    
    product\_description: str; // A detailed description of the product to make it more understandable
    
    product\_recommendation\_reason: List[str]; // The reasons that this product is recommended to the user
    
    rank: int; // The rank of the product in the recommendation, 1 being the most important
    
    index: int; // The original index of the product in the candidates list
    
\}

interface Recommendation \{

    title: str; // Main topic of the recommendation
    
    summary: str; // A summary that previews the most important points of the recommendation
    
    products: List[Product]; // The products that are recommended to the user, ranked by their importance
    
\}

End schema.
\end{prompt}

\subsection{Retrieval Database}\label{ap-retrieval-database}

As discussed earlier, each interaction or candidate in \benchf contains rich auxiliary information. We organize this information into a retrieval database, which agents access via a retrieval tool.

Unlike conventional Retrieval-Augmented Generation (RAG) setups that partition content into fixed-length chunks and retrieve them via embedding similarity, such chunking would fragment individual interactions and degrade semantic coherence in our setting. We therefore adopt a document-level retrieval strategy, where the atomic retrieval unit corresponds to a single interaction. Each document aggregates all information associated with one interaction and is organized using the following format:
\apebKeepH{The example follows the same semantic-translation, entity-placeholder, and price-perturbation convention.}

\begin{prompt}
\# Detail information of one candidate product

Product name:

\apebKeepH{[BRAND-B] [SERIES-B] Complete Vitamin C Series - Panthenol Anti-Pollution Radiant Creamy Wash, Energizing Whip Foam, Mousse Moisturizer, DD Cream Light/Natural, Face Mist, 10\% Ferulic Acid Radiant Charge Serum - Skincare}

Price:

\apebKeepH{2.3 usd}

Product description:

\apebKeepH{\texttt{<p>}[BRAND-B] [SERIES-B] Vitamin C Series\texttt{</p>}}

\apebKeepH{\texttt{<p>}1. [BRAND-B] [SERIES-B] Vitamin C + Panthenol Anti-Pollution Radiant Creamy Wash (50 mL/100 mL)\texttt{</p>}}

\apebKeepH{\texttt{<p>}Helps maintain the acid mantle as the skin barrier's first line of defense. Contains vitamin C and panthenol to brighten the skin and protect it from pollution. (...)\texttt{</p>}}

Categories:

Beauty \& Personal Care, Skincare, Serums \& Essences, Serums \& Essences

Brand name:

\apebKeepH{[BRAND-B]}

\apebKeepH{Shop name: [SHOP-B]}

Attributes:

\apebKeepH{\{'Volume': ['55 ml'], 'Manufacturing Country': ['ID'], 'Contains Dangerous Goods?': ['No'], 'Variant': ['Creamy Wash 100', 'Creamy Wash 50', 'DDCream Light', 'DDCream Natural', 'Face Mist', 'Gel Cleanser', 'Gel Moisturizer', 'Mousse Moist', 'Serum', 'Whip Foam 100', 'Whip Foam 50'], 'Edition': ['Regular Edition'], 'Region of Origin': ['ID'], 'Contains Alcohol or Aerosol': ['Contains Neither'], 'Quantity Per Pack': ['1'], 'Is Suit': ['is\_suit']\}}
\end{prompt}

The document format is adapted to its source type (product interaction, media interaction, or candidate). For example, interaction documents additionally include the corresponding action type.

We also consider the suitability of standard embedding-based retrieval for this setting. Empirically, we observe that agents primarily issue title-based queries when invoking the retrieval tool, using item titles to locate detailed information. Consequently, we adopt BM25---a robust lexical retrieval method---as the backend of the retrieval tool, rather than embedding-based similarity search.

\subsection{Rubric Details}\label{ap-experiment-rubric}

\subsubsection{\apebKeepA{Diagnostic goal proxy summarization}}\label{ap-experiment-rubric-goal}
\apebKeepA{We infer a behavior-grounded diagnostic goal proxy $g$ at the session level from all interactions in Eq.~\ref{eq:session}, including issued queries, long-viewed products ($\geq$15s), and the purchased item, using GPT-4.1.}
The model is prompted to reconstruct the latent user intent that best explains the complete search--purchase trajectory.  
\apebKeepA{Specifically, GPT-4.1 outputs a concise goal proxy $g$ that minimally and coherently aligns the observed queries with the final purchase.}
\apebKeepA{This proxy is used only as a reference for rubric diagnosis and does not determine Hit@K labels, which are computed from observed purchases in the ranked candidate lists.}

\subsubsection{Rubric perspectives}\label{ap-experiment-rubric-aspect}
\apebKeepH{Table~\ref{ap-tab:rubric-target-map} follows the four-stage organization in Section~\ref{sec:exp-why-fail} and summarizes the evaluated output and criterion for each stage; detailed scoring rules follow below.}
\apebKeepA{Diagnostic scores reported in the main analysis are averaged over Gemini-2.5-Pro, GPT-5.2, and GPT-4.1 judges; judge-sensitivity details are provided in Appendix~\ref{ap-experiment-judge-robustness}.}

\begin{table}[H]
\centering
\small
\setlength{\tabcolsep}{3pt}
\renewcommand{\arraystretch}{1.03}
\begin{tabularx}{\linewidth}{@{}>{\raggedright\arraybackslash}p{0.29\linewidth}>{\raggedright\arraybackslash}p{0.22\linewidth}>{\raggedright\arraybackslash}X@{}}
\toprule
\textbf{\apebKeepH{Stage}} & \textbf{\apebKeepH{Evaluated output}} & \textbf{\apebKeepH{Criterion and metric}} \\
\midrule
\apebKeepH{Query inference} & \apebKeepH{Inferred target $\tau$} & \apebKeepH{QI: query understanding and identification of the intended candidate category.} \\
\midrule
\apebKeepH{Historical preference reasoning} & \apebKeepH{Preferences $\mathcal{F}$} & \apebKeepH{PQ: history-grounded summary of the query-specific aspects the user cares about; PR$\downarrow$/PC: first rank/count supporting the purchase.} \\
\midrule
\apebKeepH{Recommendation quality} & \apebKeepH{Items and rationales $\mathcal{R}^{\star},\rho$} & \apebKeepH{RQ: whether the recommendations address the query and align with the user's preferences.} \\
\midrule
\apebKeepH{Agent reasoning} & \apebKeepH{ReAct steps, queries, and evidence} & \apebKeepH{AC: step coherence; QC/QH: candidate/history retrieval utility; QQ: query count.} \\
\bottomrule
\end{tabularx}
\caption{\apebKeepH{Four-stage diagnostic rubric for APeB outputs. QI/PQ/RQ/AC/QC/QH are judged scores; PR$\downarrow$ is the rank of the first purchase-supporting preference, while PC/QQ count supporting preferences/search queries.}}
\label{ap-tab:rubric-target-map}
\end{table}

\begin{itemize}
    \item \emph{Query Inference (QI)} \apebKeepH{evaluates whether the inferred target $\tau$ correctly interprets the query and identifies its intended candidate category. The judge scores how well the candidate category expressed by \texttt{target\_product} matches $(\texttt{user\_query}, \texttt{user\_goal})$ on a $[0,100]$ scale from contradiction to direct satisfaction; the resulting \texttt{query\_intent\_\allowbreak to\_target\_\allowbreak relation\_score} is the QI metric.}

    \item \emph{\apebKeepH{Historical Preference Reasoning (PQ/PR/PC)}}
\begin{itemize}
    \item \emph{Preference Quality (PQ).}
    \apebKeepH{PQ assesses whether the extracted preferences $\mathcal{F}$ faithfully and concisely summarize the query-specific aspects the user cares about, with each aspect supported by relevant history.}

    \item \emph{Preference Salience Statistics.}
    We quantify the strength of historical grounding using two statistics over validated preferences:
    \apebKeepA{\emph{Preference Minimum Rank (PR)}}, defined as the rank \apebKeepK{\(\pi_i\)} of the first preference supporting the \apebKeepA{purchased} item (lower is better), and
    \emph{preference count (PC)}, defined as the total number of distinct preferences that plausibly support the \apebKeepA{purchased} item.
    Both are computed under strict gating rules that heavily penalize irrelevant, contradictory, or ungrounded preferences.
\end{itemize}

    \item \emph{Recommendation Quality (RQ)}
    \apebKeepH{evaluates whether the recommended items $\mathcal{R}^\star$ and their rationales $\rho$ address the user's query and align with the extracted preferences $\mathcal{F}$. The judge scores item relevance and rationale support, with factual consistency checked against candidate metadata; these components are aggregated into RQ.}

    \item \emph{\apebKeepH{Agent Reasoning (AR)}}
        \begin{itemize}
            \item \emph{Agent Coherence (AC)}
            \apebKeepB{evaluates} the \emph{coherence} of each reasoning step $Re_i$ by assessing whether the agent's explicit reasoning logically justifies its stop-or-continue decision under a ReAct-style retrieval framework.
            The evaluator scores the reasoning on intent understanding, task relevance, alignment with the \apebKeepA{purchased} candidate, decision correctness, and logical completeness, penalizing unjustified leaps, hallucinated assumptions, or premature termination.
            The resulting AC score reflects the decision validity of the agent's stepwise reasoning.

            \item \emph{Candidate Query (QC)}
            \apebKeepB{evaluates} the \emph{quality} of each retrieved candidate by scoring the alignment between the issued search keywords, the agent's reasoning, and the retrieved item.
            \apebKeepB{
            We measure query relevance by scoring the retrieved evidence against the issued query. 
            Query utility measures whether the keywords accurately reflect the user query and diagnostic goal proxy, are logically derived from the ReAct reasoning, and effectively retrieve candidates that are consistent with the observed purchased product.}
            The resulting QC score captures how informative and decision-relevant the retrieval step is, penalizing generic, accidental, or misaligned candidate retrievals.

            \item \emph{History Query (QH)}
            \apebKeepB{evaluates} the \emph{quality} of retrieval history queries by scoring how effectively the issued search keywords retrieve decision-relevant user history.
            \apebKeepB{
            We measure query relevance by scoring the retrieved evidence against the issued query.
            Query utility measures whether the issued keywords retrieve history entries that provide meaningful signals for identifying the observed purchased candidate.
            }
            The resulting QH score reflects the usefulness of retrieved history for personalization, penalizing generic, accidental, or weakly informative history retrievals.

            \item \emph{Query Quantity (QQ)}
            measures the total number of retrieval queries issued by the agent, defined as the sum of candidate queries and history queries, capturing the overall extent of retrieval tool usage during reasoning.

        \end{itemize}
\end{itemize}
\apebKeepA{For the ReAct failure-source analysis in Appendix~\ref{ap-react-failure-analysis}, AC is averaged at the reasoning-step or round level, QC/QH summarize the quality of retrieved candidate/history evidence, and QQ is summarized through total query count across candidate/history \apebKeepB{across multiple steps}.}

\subsubsection{\apebKeepH{Judge Robustness}}
\label{ap-experiment-judge-robustness}
\begin{table}[H]
\centering
\footnotesize
\setlength{\tabcolsep}{3pt}
\renewcommand{\arraystretch}{1.05}
\begin{tabularx}{\linewidth}{@{}>{\raggedright\arraybackslash}Xrrrr@{}}
\toprule
\textbf{\apebKeepH{Model}} & \textbf{\apebKeepH{Gemini}} & \textbf{\apebKeepH{GPT-5.2}} & \textbf{\apebKeepH{GPT-4.1}} & \textbf{\apebKeepH{Mean}} \\
\midrule
GPT-4o & 68.9 & 69.1 & 69.2 & 69.1 \\
GPT-4.1-mini & 64.1 & 64.1 & 64.2 & 64.1 \\
Qwen3 & 67.3 & 67.2 & 67.3 & 67.3 \\
\bottomrule
\end{tabularx}
\caption{\apebKeepH{Judge sensitivity of mean diagnostic-rubric scores on 500 randomly sampled \smartqB sessions. Columns denote Gemini-2.5-Pro, GPT-5.2, and GPT-4.1 judges.}}
\label{ap-tab:judge-sensitivity}
\end{table}

\apebKeepH{Table~\ref{ap-tab:judge-sensitivity} shows nearly identical relative rankings across the three judges; human pairwise preference checks agree with the rubric preference in 96 of 100 cases.}
\apebKeepB{This supports using rubric scores as diagnostic evidence while preserving Hit@K as the task-native evaluation metric.}

\subsubsection{Rubric Validity via Correlation with Task Metrics}
\label{ap-experiment-rubric-verification}
\apebKeepB{The complementary judge-sensitivity evidence is summarized in Appendix~\ref{ap-experiment-judge-robustness}; this subsection focuses on rubric--Hit@K alignment.}
To evaluate output quality at the component level, we adopt an \emph{LLM-as-a-Judge} paradigm to score model behaviors along multiple rubric dimensions (Appendix~\ref{ap-experiment-rubric}). 
Prior work on LLM-based evaluation typically establishes validity through agreement with human annotations or inter-judge consistency~\cite{judge-verify-1,judge-verify-2}. 
While such criteria assess annotator reliability, they do not directly verify whether rubric scores faithfully reflect task success.

\begin{figure*}[!t]
    \centering
    \begin{subfigure}[t]{0.31\textwidth}
        \centering
        \includegraphics[width=\linewidth]{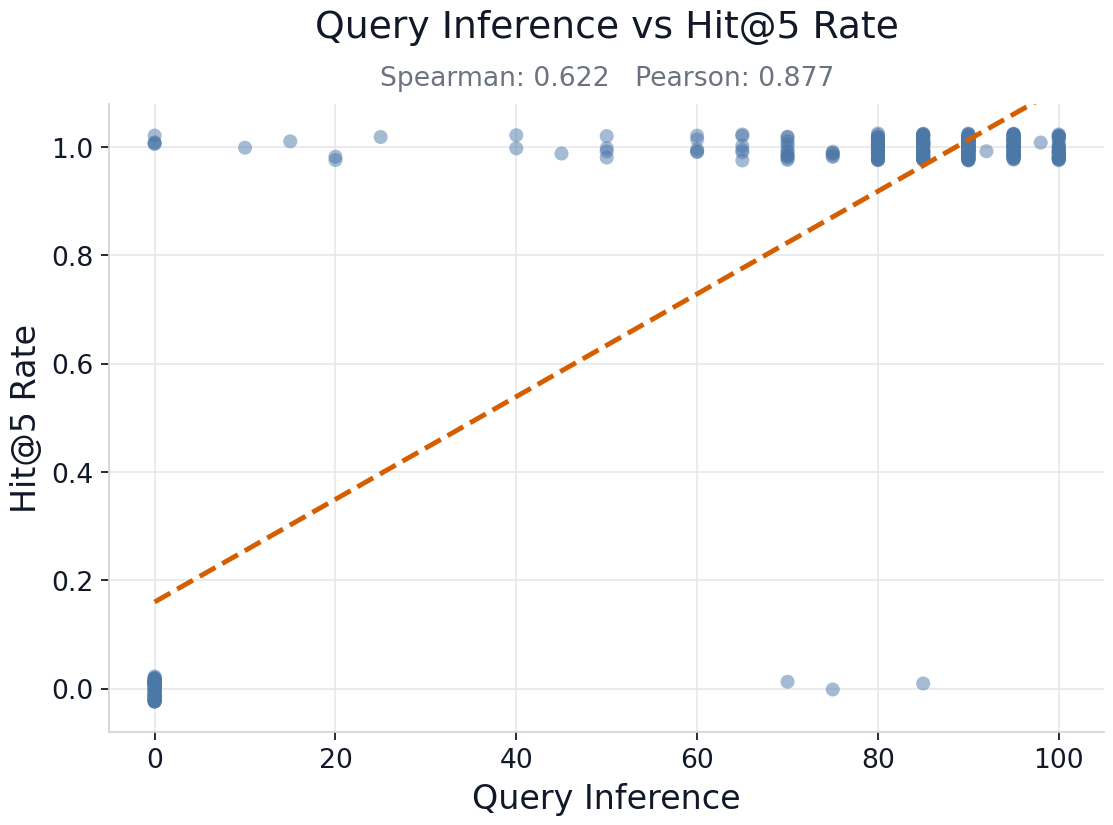}
        \caption{QI vs. Hit@5}
    \end{subfigure}
    \hfill
    \begin{subfigure}[t]{0.31\textwidth}
        \centering
        \includegraphics[width=\linewidth]{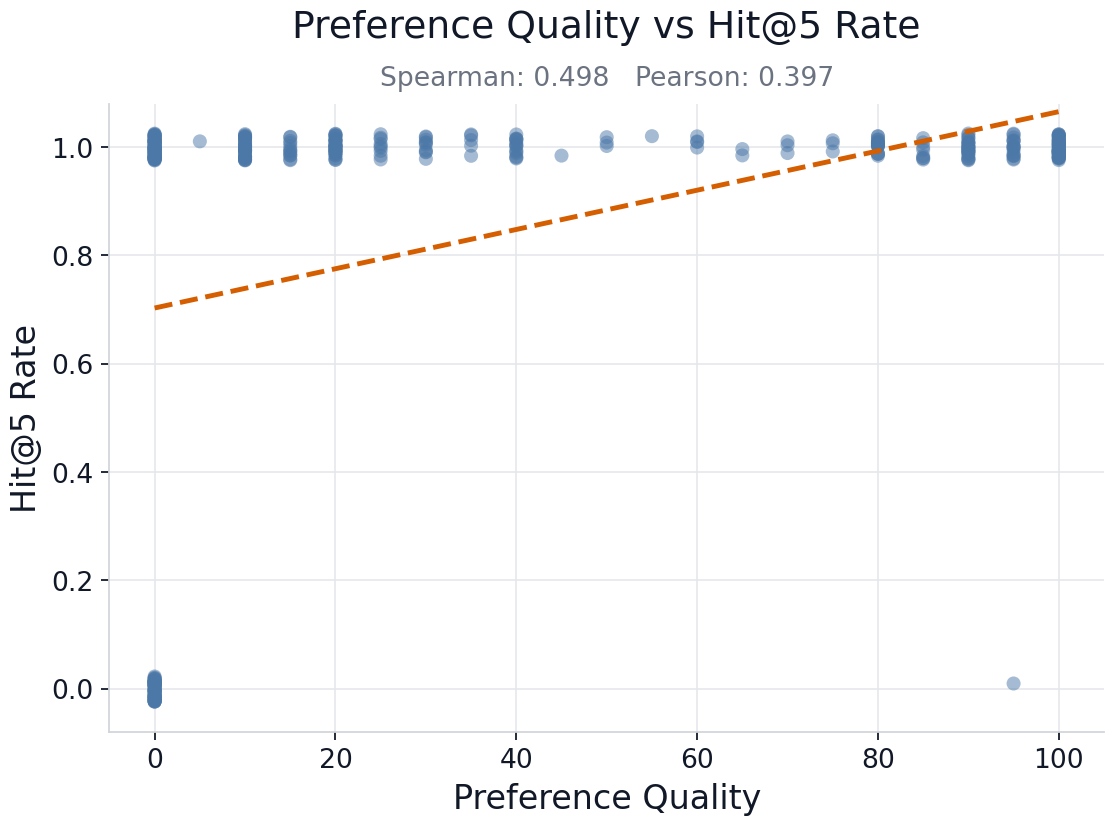}
        \caption{PQ vs. Hit@5}
    \end{subfigure}
    \hfill
    \begin{subfigure}[t]{0.31\textwidth}
        \centering
        \includegraphics[width=\linewidth]{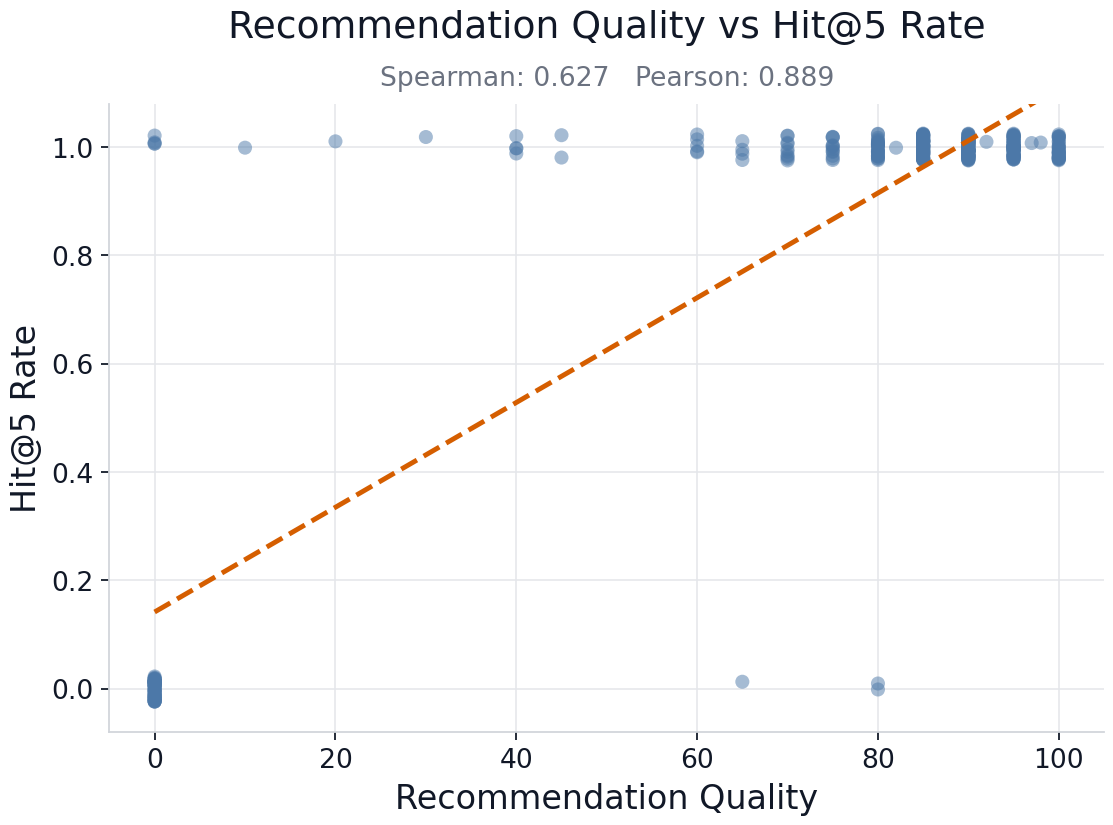}
        \caption{RQ vs. Hit@5}
    \end{subfigure}

    \vspace{0.35em}

    \begin{subfigure}[t]{0.31\textwidth}
        \centering
        \includegraphics[width=\linewidth]{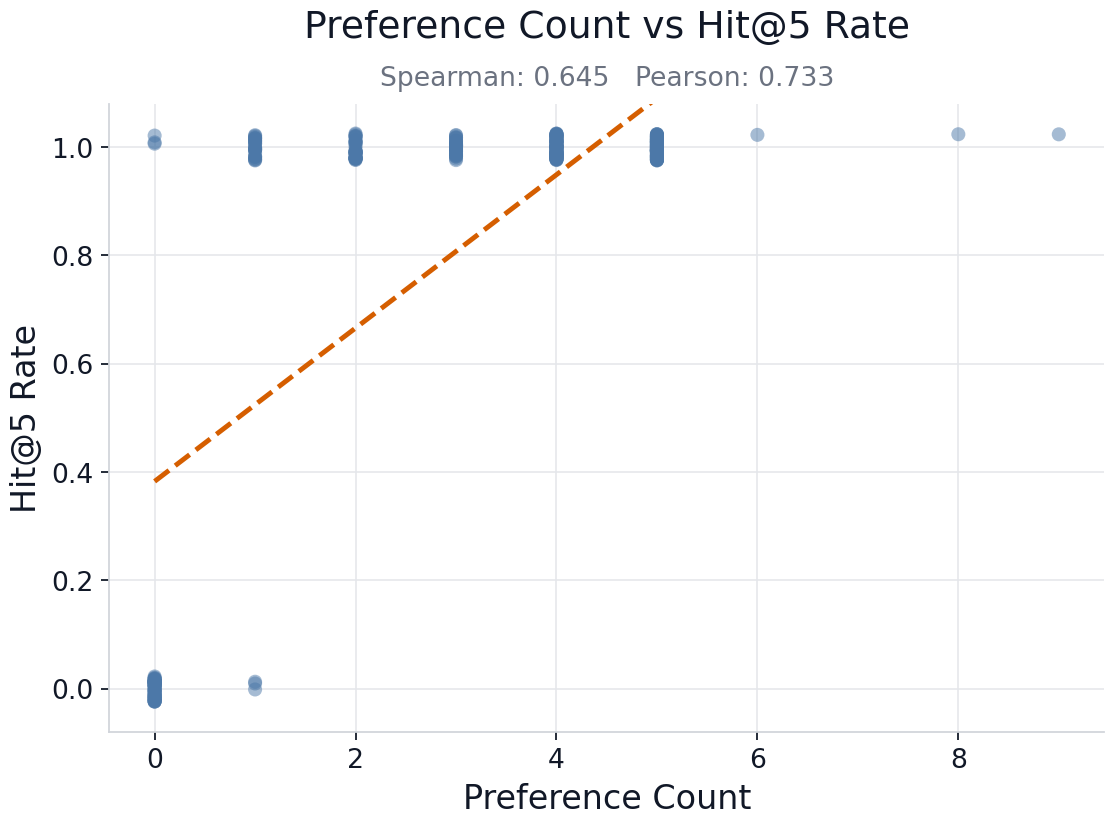}
        \caption{PC vs. Hit@5}
    \end{subfigure}
    \hfill
    \begin{subfigure}[t]{0.31\textwidth}
        \centering
        \includegraphics[width=\linewidth]{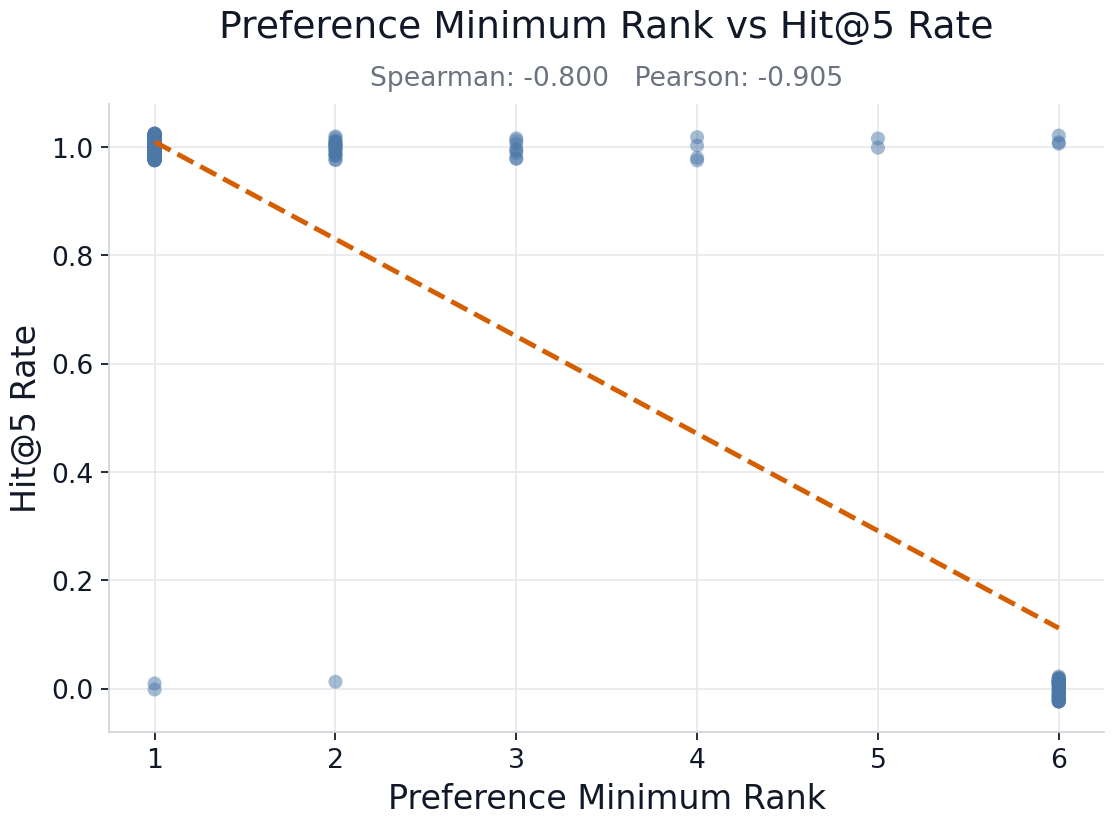}
        \caption{PR vs. Hit@5 ($\downarrow$)}
    \end{subfigure}
    \hfill
    \begin{subfigure}[t]{0.31\textwidth}
        \centering
        \includegraphics[width=\linewidth]{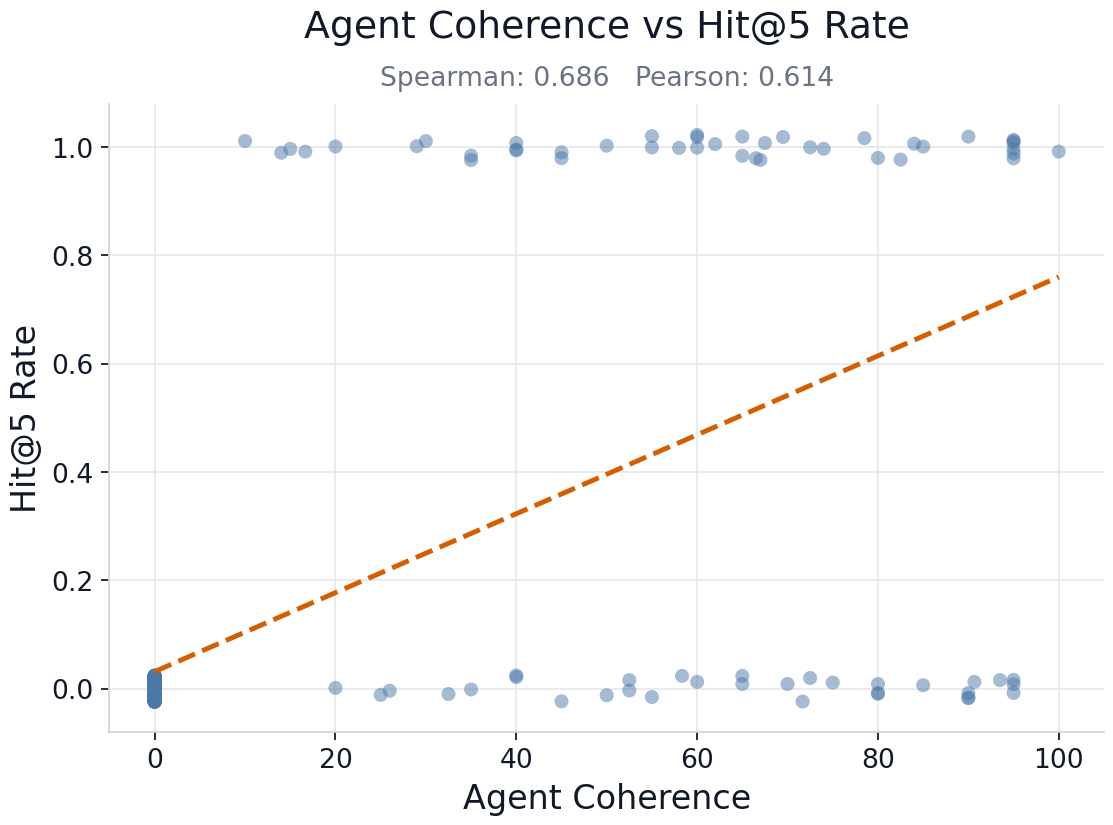}
        \caption{AC vs. Hit@5}
    \end{subfigure}
    \caption{Session-level alignment between diagnostic rubric scores and task-native Hit@5. Panels cover LLM diagnostics (QI, PQ, RQ, PC, PR) and ReAct agent coherence (AC); PR is a rank metric, so lower values indicate stronger preference salience.}
    \label{ap-fig:rubric-hitk-correlation}
\end{figure*}

Following recent work on metric validity~\cite{corr-verify-1,corr-verify-2}, we instead \apebKeepA{assess} our rubrics by examining their alignment with \emph{task-native success signals}. 
Specifically, we use Hit@K as the reference metric family and analyze the plotted Hit@5 alignment between rubric-based scores and task success across models and agentic workflows (Figure~\ref{ap-fig:rubric-hitk-correlation}).
\apebKeepD{Our key assumption is that, if subtask-level rubric scores produced by a Judger LLM exhibit strong correlation with Hit@K within each model or experimental setting, these scores can be treated as \apebKeepA{supporting diagnostic evidence} for diagnosing component-level reasoning failures.}

\begin{figure*}[!t]
    \centering
    \begin{subfigure}[t]{0.47\textwidth}
        \centering
        \includegraphics[width=\linewidth]{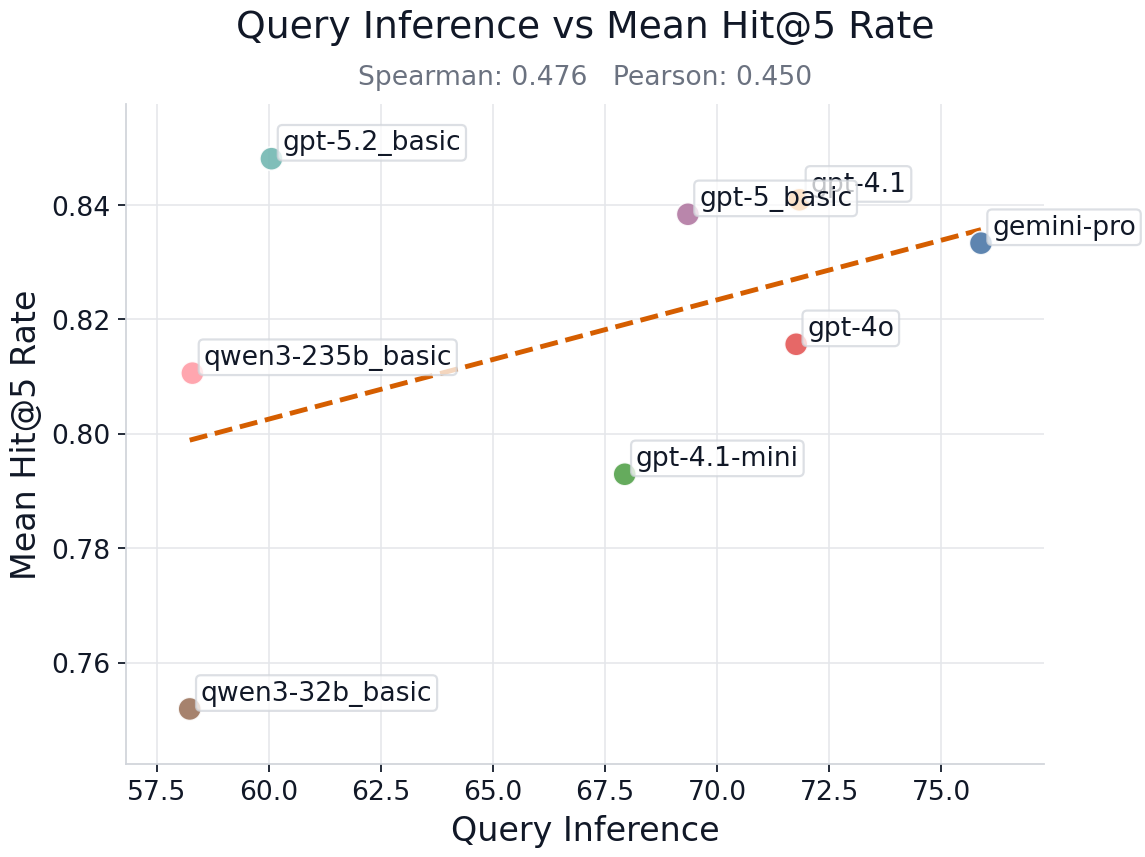}
        \caption{Mean QI vs. mean Hit@5}
    \end{subfigure}
    \hfill
    \begin{subfigure}[t]{0.47\textwidth}
        \centering
        \includegraphics[width=\linewidth]{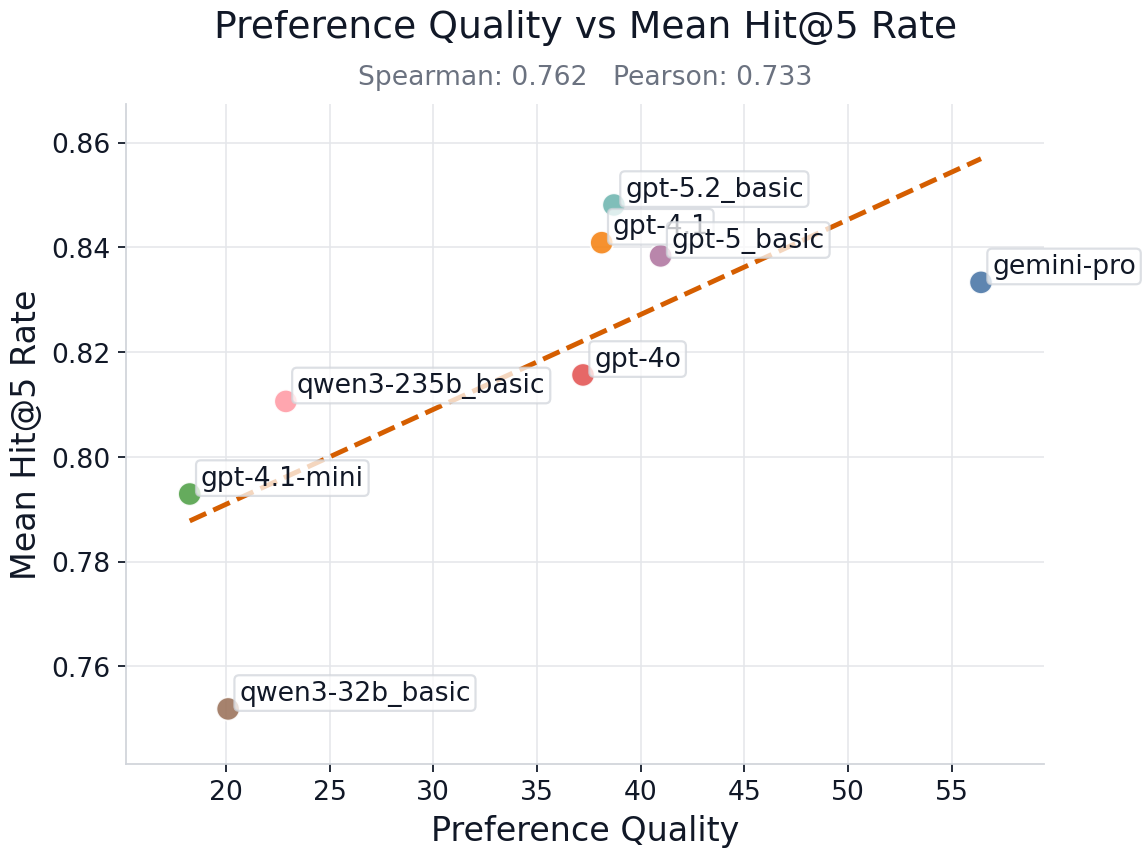}
        \caption{Mean PQ vs. mean Hit@5}
    \end{subfigure}

    \vspace{0.35em}

    \begin{subfigure}[t]{0.47\textwidth}
        \centering
        \includegraphics[width=\linewidth]{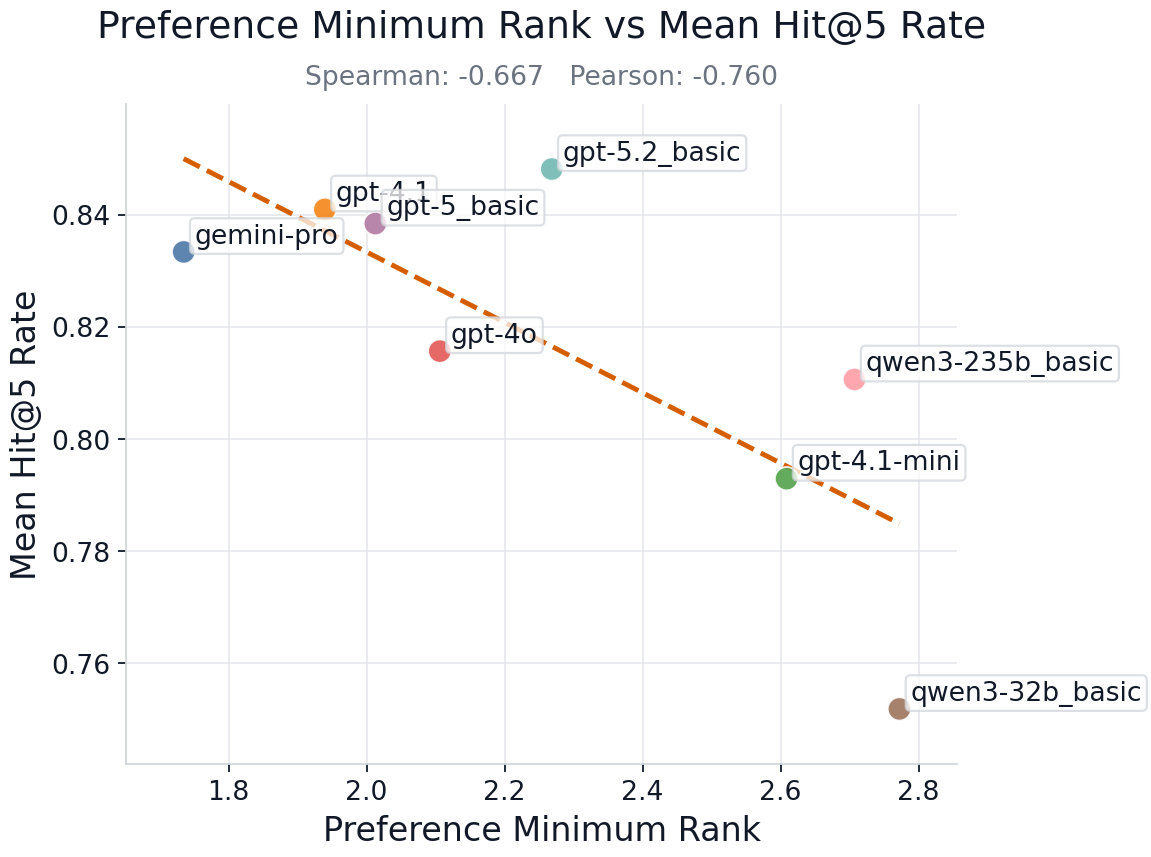}
        \caption{Mean PR vs. mean Hit@5 ($\downarrow$)}
    \end{subfigure}
    \hfill
    \begin{subfigure}[t]{0.47\textwidth}
        \centering
        \includegraphics[width=\linewidth]{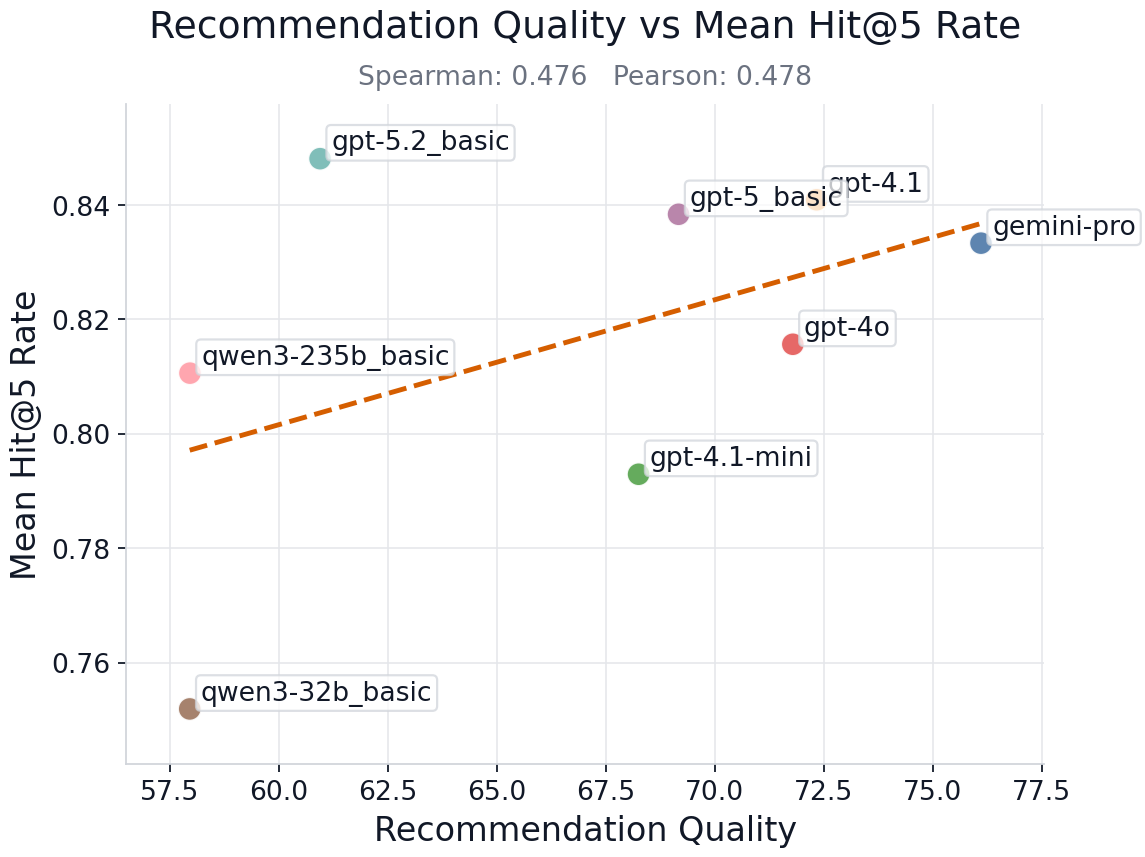}
        \caption{Mean RQ vs. mean Hit@5}
    \end{subfigure}

    \caption{Model-level rubric--Hit@5 alignment under the \smartq hard-candidate setting. Each point is one model backbone; panels aggregate the diagnostics used in Figure~\ref{fig:exp-rubric}, showing whether cross-model task performance tracks rubric aspects.}
    \label{ap-fig:rubric-model-comparison}
\end{figure*}

Beyond correlation strength, we further examine whether rubric scores preserve \emph{relative performance differences} across models and workflows.
Figure~\ref{ap-fig:rubric-model-comparison} compares model-level mean rubric scores against mean Hit@5 from multiple rubric perspectives, showing that models with higher task success consistently achieve stronger rubric scores along corresponding dimensions.
This consistency supports the use of rubric-based evaluation not only as a diagnostic signal within a model, but also as a meaningful basis for cross-model and cross-workflow comparison.

\apebKeepA{Together, these results support using rubric degradation as diagnostic evidence for task-level failures, while preserving Hit@K as the primary task-native outcome.}

\subsection{\apebKeepH{Summary of LLM Roles and Audits}}
\label{ap-experiment-llm-role-audit}
\apebKeepH{This subsection consolidates the LLM roles and audit evidence reported across the main text and appendix. Table~\ref{ap-tab:llm-role-audit} serves as a cross-reference: each row identifies one role, summarizes its principal check, and points to the section containing the corresponding setup and results.}

\begin{table}[H]
\centering
\footnotesize
\setlength{\tabcolsep}{2.5pt}
\renewcommand{\arraystretch}{1.05}
\begin{tabularx}{\linewidth}{@{}>{\raggedright\arraybackslash}p{0.23\linewidth}>{\raggedright\arraybackslash}X>{\centering\arraybackslash}p{0.19\linewidth}@{}}
\toprule
\textbf{\apebKeepH{LLM use}} & \textbf{\apebKeepH{Role and principal check}} & \textbf{\apebKeepH{Detailed in}} \\
\midrule
\apebKeepH{Query--view--purchase alignment} & \apebKeepH{Construction filter; human and cross-model alignment audit ($N=100$).} & \apebKeepH{App.~\ref{ap-benchmark-filter-session-rule3}} \\
\midrule
\apebKeepH{Intent/refined-query localization} & \apebKeepH{Selects \smartq/\matchq indices; human exact-pair audit ($N=100$).} & \apebKeepH{App.~\ref{ap-benchmark-filter-session-distinct-query}} \\
\midrule
\apebKeepH{Purchase validation} & \apebKeepH{Pairwise recoverability diagnostic with query, history, and metadata controls ($N=500$).} & \apebKeepH{App.~\ref{ap-hard-candidate-controls}} \\
\midrule
\apebKeepH{Diagnostic goal proxy} & \apebKeepH{Session-derived reference for the component-level diagnostic rubric.} & \apebKeepH{App.~\ref{ap-experiment-rubric-goal}} \\
\midrule
\apebKeepH{Rubric judging} & \apebKeepH{Scores diagnostic outputs; multi-judge sensitivity and human-preference audit.} & \apebKeepH{App.~\ref{ap-experiment-judge-robustness}} \\
\midrule
\apebKeepH{VQRA rewriting} & \apebKeepH{History-conditioned query refinement evaluated through Hit@K and rubric analysis.} & \apebKeepH{Sec.~\ref{sec:exp-rq3-vqra}} \\
\bottomrule
\end{tabularx}
\caption{\apebKeepH{Cross-reference summary of LLM roles in APeB construction, diagnostics, and evaluation. Each row gives the role, its principal check, and the section containing the full setup and evidence.}}
\label{ap-tab:llm-role-audit}
\end{table}

\raggedbottom

\subsection{\ouragent Detail}\label{ap-experiment-VQRA}

\subsubsection{\texorpdfstring{\apebKeepH{Self-Rewrite Experiment}}{Self-Rewrite Experiment}}
\label{ap-vqra-self-rewrite-experiment}

\apebKeepA{Section~\ref{sec:exp-why-fail} and Figure~\ref{fig:exp-rubric} attribute the performance drop on \smartq queries primarily to weak query inference. To empirically verify this hypothesis, we construct a query-refinement pipeline.}
\apebKeepA{Given an \smartq query, which is intentionally under-specified and weakly aligned with the final purchased product, we refine it into an explicit \matchq query context using selected user interaction history. Selected same-category context follows the intent-query category-prior convention in Appendix~\ref{ap-experiment-llm-selection-history}.}
\apebKeepA{Concretely, in the main rewrite setting, we select $k=20$ product and media interaction records from the selected same-category history context as evidence. 
A shared backbone (GPT-4.1) rewrites the original \smartq query into a precise and discriminative keyword query, and the rewritten query is then used as the user query for personalized product retrieval. More than 92\% of evaluated sessions have enough same-category histories for this context.}
\apebKeepA{We additionally evaluate a self-rewrite variant, where the evaluated backbone rewrites its own query; VQRA(GPT-4.1) uses GPT-4.1 as a shared rewrite module.}

\begin{table}[H]
\centering
\small
\setlength{\tabcolsep}{4pt}
\begin{tabular}{lccc}
\toprule
\textbf{Setting} & \textbf{Gemini} & \textbf{\apebKeepH{GPT-5.2}} & \textbf{\apebKeepH{GPT-5.2T}} \\
\midrule
Vanilla & 22.1 & 25.6 & 24.2 \\
VQRA(self) & 25.4 & 26.5 & \textbf{26.6} \\
VQRA(GPT-4.1) & 25.2 & 26.4 & \textbf{26.9} \\
\bottomrule
\end{tabular}
\caption{Self-rewrite ablation for VQRA on \smartq hard-candidate cases. GPT-5.2T denotes GPT-5.2Think; both VQRA variants use the same intent-category history context.}
\label{ap-tab:vqra-self-rewrite}
\end{table}

\apebKeepA{Self-rewrite improves Hit@1 for all three reported backbones and is comparable to the GPT-4.1 rewrite variant, indicating that VQRA's gains mostly come from how the model utilizes history, not solely due to GPT-4.1 acting as an external assistant.}
\apebKeepB{By substituting the original \smartq with its refined counterpart while keeping the retrieval pipeline fixed, this design directly tests whether the model has potential to resolve query ambiguity through pipeline redesign and whether doing so can improve performance.}

\subsubsection{\texorpdfstring{\apebKeepH{Qualitative Query-Refinement Example}}{Qualitative Query-Refinement Example}}
\label{ap-vqra-example}
\apebKeepH{In an anonymized session, the initial query ``cloves, rosemary for haircare'' leaves the intended product type underspecified. VQRA uses the user's history to rewrite it as ``rosemary hair growth oil'' and selects the type of rosemary hair growth oil subsequently purchased by the user. Methods following the literal wording rank raw rosemary and clove ingredients.}

\end{document}